\documentclass[11pt,a4paper]{article}
\usepackage{color}
\usepackage[margin=1in]{geometry}
\usepackage{fullpage,mathrsfs,graphicx,framed,color,amssymb,amsmath,amsthm}
\usepackage[boxed, noline, ruled, linesnumbered]{algorithm2e}
\usepackage{epsfig}
\usepackage{latexsym,amsfonts,amsbsy,amssymb}

\usepackage{lmodern}

\usepackage{enumerate}
\usepackage{tikz}
\usepackage{stmaryrd} 
\usepackage{pgfplots}
\pgfplotsset{compat=1.17}
\usepackage{multirow}
\usepackage{booktabs}
\usepackage{arydshln} 
\usepackage{bbding} 

\usepackage{subcaption}
\usepackage{changes} % for revision

\usepackage[utf8]{inputenc} % ^T

\usepackage{caption}
\usepackage[colorlinks=true]{hyperref}
\usepackage[capitalize, sort, compress]{cleveref}

\crefname{equation}{}{}
\crefname{theorem}{Theorem}{Theorems}
\crefname{figure}{Figure}{Figures}
\crefname{table}{Table}{Tables}
\crefname{section}{Section}{Sections}
\crefname{example}{Example}{Examples}
\crefname{algocf}{Algorithm}{Algorithms}
\crefname{lemma}{Lemma}{Lemmas}
\crefname{remark}{Remark}{Remarks}

\makeatletter %'@' is now a normal "letter" for TeX

\@addtoreset{equation}{section}
\makeatother %'@' is restored as a "non-letter" character for TeX
\allowdisplaybreaks % For long formula
\newtheorem{theorem}{Theorem}[section]

\newtheorem{remark}{Remark}[section]

\allowdisplaybreaks 
\numberwithin{equation}{section}

\begin{document}
\title{fTNN: a tensor neural network for fractional PDEs}

\author{Qingkui Ma\footnote{School of Mathematics and Statistics \& 
Hubei Key Laboratory of Mathematical Sciences,
Central China Normal University, Wuhan 430079, China
(maqingkui@mails.ccnu.edu.cn).}\ \ \ 
Hehu Xie\footnote{SKLMS, NCMIS, Academy of Mathematics and Systems Science,
Chinese Academy of Sciences, No.55, Zhongguancun Donglu, Beijing 100190, 
China, and School of Mathematical Sciences, University of Chinese Academy
of Sciences, Beijing, 100049, China (hhxie@lsec.cc.ac.cn).}\ \ \ and \ \ 
Xiaobo Yin\footnote{School of Mathematics and Statistics \& Key Laboratory of Nonlinear Analysis \& Applications (Ministry of Education),
Central China Normal University, Wuhan 430079, China (yinxb@ccnu.edu.cn).}}
\date{}
\maketitle
%==============================================================================================
\begin{abstract}
We develop the fTNN, a deterministic tensor neural network subspace method for problems involving the fractional Laplacian on bounded domains, taking the fractional Poisson equation and time-dependent fractional advection-diffusion equation as typical representatives. The work employs a geometry-adapted integration split featuring a spatially dependent near-field radius, which decomposes the fractional Laplacian into three contributions: a singular near field, a regular interior far field, and an analytical exterior far field. Then the singular radial integrals are treated by Gauss-Jacobi quadrature, the regular radial integrals by Gauss quadrature, and the angular variables by deterministic angular quadrature, yielding a fully deterministic integration framework of the fractional Laplacian operator.
To accurately resolve low-regularity solutions and the associated loss functional, we construct boundary-singularity-aware trial functions enriched with explicit boundary features, and propose two  strategies for automatically selecting the leading exponent and evaluating the loss function from the singularity structure induced by the fractional operator, or jointly by the fractional operator and the source term. For time-dependent fractional PDEs, we design a spatiotemporally separable neural network that factorizes the time-space residual into a sum of low-dimensional temporal and spatial integrals, and we integrate this representation with an alternating neural network subspace optimization strategy for efficient training.
Numerical experiments show that the proposed framework attains high accuracy on the tested benchmarks and improves substantially over existing fPINN and Monte Carlo baselines, particularly for problems with strong boundary singularities and long-time simulations.
	
	
\end{abstract}

\section{Introduction}
Fractional partial differential equations involving the fractional Laplacian have attracted sustained attention in analysis, scientific computing, and applications because they arise naturally in anomalous transport, long-range interaction, and nonlocal diffusion models; see, for example, \cite{Bonito2017,Caffarelli2006,DElia2020,Lischke2018}. Among these, models involving the spatial fractional Laplacian are substantially more challenging to treat numerically than those involving only time-fractional derivatives, especially on bounded domains \cite{Sheng2024}.
The main difficulties stem from the simultaneous presence of hypersingular nonlocal kernels, exterior Dirichlet constraints, and reduced boundary regularity of solutions \cite{pang2015,ros2016boundary,sheng2023efficient}. These features make the design of accurate and robust numerical methods for fractional Laplacian problems on bounded domains especially demanding.

In this work, we first consider the fractional Poisson equation (fPE) with homogeneous Dirichlet exterior conditions:
\begin{equation}\label{eq:fra_lapla_problem}
\left\{
\begin{aligned}
(-\Delta)^{\alpha / 2} u(\boldsymbol{x}) &= f(\boldsymbol{x}), &&\boldsymbol{x} \in \Omega, \\
u(\boldsymbol{x}) &= 0, &&\boldsymbol{x} \in \Omega^c.
\end{aligned}
\right.
\end{equation}
Here \(\Omega\subset\mathbb{R}^d\) is either a hyperrectangle or the unit ball, and \(0<\alpha<2\). The fractional Laplacian operator $(-\Delta)^{\alpha / 2}$ admits several non-equivalent definitions in different settings \cite{Lischke2018}. In this paper, we adopt the Riesz definition with zero exterior extension:
\begin{equation}\label{eq:def_lap}
(-\Delta)^{\alpha/ 2} u(\boldsymbol{x}) = C_{d,\alpha} \int_{\mathbb{R}^d} \frac{u(\boldsymbol{x}) - u(\boldsymbol{y})}{\| \boldsymbol{y} - \boldsymbol{x} \|^{d+\alpha}} d\boldsymbol{y} \quad\mbox{with}\quad C_{d,\alpha}=\frac{2^{\alpha -1}\alpha\Gamma \left(\alpha/ 2+d/2 \right)}{\pi^{d/2} \Gamma (1-\alpha/ 2)}.
\end{equation}
We also consider the following time-dependent fractional partial differential equation (fPDE) posed on \(\Omega\) with zero exterior condition \cite{pang2019fpinns}:
\begin{equation}\label{eq:fPDE}
\left\{
\begin{aligned}
L_{t,\boldsymbol{x}}\{u(\boldsymbol{x},t)\}=\frac{\partial^\gamma u(\boldsymbol{x},t)}{\partial t^\gamma}+L_{\boldsymbol{x}}\{u(\boldsymbol{x},t)\}
&=f(\boldsymbol{x},t),
\quad \boldsymbol{x}\in\Omega,\ t\in(0,T],\\
u(\boldsymbol{x},t)&=0,
\qquad\quad\:\: \boldsymbol{x}\in\Omega^c,\ t\in(0,T],\\
u(\boldsymbol{x},0)&=u_0(\boldsymbol{x}),
\quad\:\: \boldsymbol{x}\in\Omega.
\end{aligned}
\right.
\end{equation}
Here \(L_{\boldsymbol{x}}\) denotes a linear spatial operator involving the fractional Laplacian and acting on \(u\). A typical example is
\begin{equation}\label{eq:space_L}
L_{\boldsymbol{x}}
= (-\Delta)^{\alpha/2}+\boldsymbol{v}\cdot\nabla,
\end{equation}
where \(\boldsymbol{v}\cdot\nabla\) is the advection term associated with a prescribed velocity field \(\boldsymbol{v}\). The time-fractional derivative in \cref{eq:fPDE} is understood in the Caputo sense:
\begin{equation}\label{eq:Caputo}
\frac{\partial^\gamma u(\boldsymbol{x},t)}{\partial t^\gamma}
=\frac{1}{\Gamma(1-\gamma)}
\int_0^t (t-\tau)^{-\gamma}
\frac{\partial u(\boldsymbol{x},\tau)}{\partial \tau}\, d\tau,
\qquad 0<\gamma\le 1.
\end{equation}
When \(\gamma=1\), the Caputo derivative reduces to the classical first-order time derivative. By choosing different values of \(\gamma\) and \(\alpha\), and by including or excluding the advection term, \cref{eq:fPDE} recovers several important classes of fractional models. In particular, when \(0<\gamma<1\), \(\alpha=2\), and \(\boldsymbol{v}=0\), it reduces to a time-fractional diffusion equation \cite{lin2025TFPIDE}, which describes subdiffusion with memory effects. When \(\gamma=1\) and \(\alpha\in(0,2)\) with \(\boldsymbol{v}=0\), it becomes a time-dependent space-fractional diffusion equation that captures nonlocal transport phenomena such as Lévy-flight-type diffusion. When \(0<\gamma<1\), \(\alpha\in(0,2)\), and \(\boldsymbol{v}\neq 0\), \cref{eq:fPDE} becomes a time-space fractional advection-diffusion equation, incorporating both temporal memory and spatial nonlocality. Such models are widely used to describe anomalous transport phenomena \cite{Lischke2018}.

Classical discretization methods, such as the finite difference method \cite{Huang2024grid,Meerschaert2006FDM,Tadjeran2007} and the finite element method \cite{Acosta2017,Ainsworth2018,Gao2019AFE,Sheng2024}, remain important tools for solving fractional PDEs. However, the nonlocal character of fractional operators typically leads to dense matrices or globally coupled discrete systems, which are expensive to assemble, store, and solve. In addition, boundary singularities reduce regularity and complicate mesh design, quadrature, and error control, particularly in three dimensions. These issues motivate the development of alternative high-accuracy mesh-free methods.

In recent years, neural network-based solvers have provided a complementary framework for PDEs. After the pioneering work of Lagaris et al. \cite{lagaris1998artificial}, physics-informed neural networks (PINNs) \cite{raissi2019physics} and their fractional extensions have been developed for a variety of forward and inverse problems. Representative examples include fPINNs \cite{pang2019fpinns}, bi-orthogonal fPINNs \cite{ma2023bi}, spectral-fPINNs \cite{zhang2025spectral}, and general Monte Carlo PINN formulations on irregular domains \cite{wang2024gmc}. 
For high-dimensional fPDEs, Monte Carlo discretization has become an important direction because it avoids the explicit construction of dense nonlocal matrices. From a probabilistic perspective, Sheng et al. \cite{sheng2023efficient} proposed an efficient Monte Carlo solver for fractional PDEs. Their approach extends the walk-on-spheres strategy by using a Feynman-Kac representation based on the Green's function for the unit ball, enabling computations in complex domains and high dimensions. Within the PINN framework, Guo et al. \cite{guo2022monte} proposed the so-called MC-fPINN method, in which the fractional Laplacian is approximated by Monte Carlo sampling after splitting the integral into integrals over a neighborhood \(B_{r_{0}}(\boldsymbol{x})\) of \(\boldsymbol{x}\) and its complement:
\begin{equation*}
(-\Delta)^{\alpha/ 2} u(\boldsymbol{x}) = C_{d,\alpha} \left( \int_{\boldsymbol{y} \in B_{r_{0}}(\boldsymbol{x})} \frac{u(\boldsymbol{x}) - u(\boldsymbol{y})}{\|\boldsymbol{x} - \boldsymbol{y}\|^{d+\alpha}} d\boldsymbol{y} + \int_{\boldsymbol{y} \notin B_{r_{0}}(\boldsymbol{x})} \frac{u(\boldsymbol{x}) - u(\boldsymbol{y})}{\|\boldsymbol{x} - \boldsymbol{y}\|^{d+\alpha}} d\boldsymbol{y} \right).
\end{equation*}
Splitting the integral into a singular near-field part and a regular far-field part enables tailored numerical treatment. Direct Monte Carlo sampling of the singular part, however, tends to introduce substantial variance and sampling error, limiting attainable accuracy and slowing convergence. Moreover, the performance of such sampling-based methods is often sensitive to the choice of splitting radius and to the way the singular component is truncated.
To reduce sampling variance and improve radial integration accuracy, Hu et al. \cite{hu2024tackling} replaced the Monte Carlo estimate of the one-dimensional singular radial integral by Gauss-Jacobi quadrature for the near-field part. Extending this strategy, we recently proposed the Quadrature-Enhanced Monte Carlo fPINN (QE-MC-fPINN) method \cite{ma2026quadrature} which adopts a geometry-adaptive decomposition of the fractional Laplacian based on a spatially varying near-field radius. Furthermore, the method combines deterministic radial quadrature with Monte Carlo angular sampling, and embeds the resulting operator approximation into a feature-enhanced neural network trial space tailored to low-regularity solutions.
Although it improves upon representative state-of-the-art methods, QE-MC-fPINN also reveals a critical bottleneck: the dominant error is no longer the radial singularity, but the stochastic noise from the angular Monte Carlo sampling. Eliminating this noise while preserving the geometry-adaptive decomposition is the primary motivation to develop a fully deterministic approach in this paper.

A second, complementary ingredient of our framework comes from recent neural network subspace and tensor neural network methods \cite{li2024tensor,liao2022solving,wang2024solving,wang2024computing,wang2024tensor}. In particular, Lin et al. \cite{lin2025TFPIDE} proposed a tensor neural network subspace method to solve time-fractional partial integro-differential equations. The method combines a power-law temporal factor $t^{\mu}$, Gauss-Jacobi quadrature, and an alternating optimization strategy. However, the multidimensional nonlocality of the fractional Laplacian poses significant challenges for reducing high-dimensional integrals to one-dimensional representations.
To address this, we propose a spatiotemporally separable neural network (STSNN). The STSNN constructs a structured solution subspace such that the PDE residual is decomposed into decoupled temporal and spatial terms, enabling efficient representation and optimization for time-dependent nonlocal problems.

Building on the MC-fPINN \cite{guo2022monte}, Improved MC-fPINN \cite{hu2024tackling}, and especially the QE-MC-fPINN method \cite{ma2026quadrature}, the present work develops a deterministic and subspace-based framework for fractional PDEs on bounded domains. We retain the geometry-adaptive near-/far-field philosophy of the QE-MC-fPINN, but replace the remaining angular Monte Carlo sampling by deterministic angular quadrature, incorporate boundary-singularity-aware trial spaces through adaptive exponent selection, and combine these ideas with STSNN and alternating subspace optimization for time-space fractional PDEs. The main contributions of this paper are summarized as follows:
\begin{enumerate}
\item A fully deterministic quadrature framework for the fractional Laplacian is established, which employs a spatially adaptive near-field radius to decompose the operator into three parts. The radial integrals are evaluated via Gauss-Jacobi or Gauss quadrature, and the angular integrals for all three parts are approximated using deterministic quadrature rules.

\item We construct boundary-singularity-aware neural-network trial spaces with explicit boundary features \(b(\boldsymbol{x})^{\mu_j}\), and propose two strategies, BFE and BRFE, to determine the leading exponent according to the singularity structure induced by the fractional operator, or jointly by the fractional operator and the source term, respectively. These two strategies supply effective ways to treat low-regularity behaviors of the exact solutions.

\item For time-space fractional PDEs, we propose a STSNN subspace method. The resulting time-space residual factorizes into sums of products of lower-dimensional temporal and spatial integrals, and this structure is combined naturally with alternating subspace optimization. Consequently, the proposed method handles both short- and long-time simulations efficiently: the separable structure allows dense temporal Gauss quadrature without memory explosion, which is particularly advantageous for the nonlocal memory of the Caputo derivative.
\end{enumerate}

The remainder of this paper is organized as follows. In \cref{Sec:ML_fPDE}, we recall Gauss-Jacobi quadrature, introduce the STSNN architecture, detail the deterministic discretization of the fractional Laplacian, and construct the corresponding loss functions. In \cref{Sec:Optimization}, we introduce the neural-network subspace procedure used to determine the trial basis and linear coefficients. Numerical experiments in one, two, and three dimensional cases are reported in \cref{Sec:Numerical} to demonstrate the accuracy and efficiency of the proposed methods.

\section{Machine learning methods for fPDEs}\label{Sec:ML_fPDE}
This section presents the analytical and computational components of the fully deterministic framework developed in this work: Gauss-Jacobi quadrature, the STSNN architecture, the deterministic discretization of the fractional Laplacian, and the associated loss constructions.

\subsection{Gauss-Jacobi quadrature}\label{sec:G_J_qua}
Gauss-Jacobi quadrature is an efficient high-precision method for approximating definite integrals. It computes integrals over $[-1,1]$ with the weight function $(1 - x)^{\beta_{1}}(1+x)^{\beta_{2}}$ using $N$ nodes and weights (see \cite{brzezinski2018computation,shen2011spectral}):
\begin{equation}\label{Def_gauss_jacobi_11}
\int_{-1}^1 (1-x)^{\beta_{1}}(1+x)^{\beta_{2}}f(x)dx = \sum_{i=1}^N \widehat{w}_{i}^{(\beta_{1},\beta_{2})} f(\widehat{x}_{i}^{(\beta_{1},\beta_{2})}) + R_N(f),
\end{equation}
with $\beta_{1}> -1$, $\beta_{2}> -1$, where $R_N(f)$ is the integration error. The quadrature formula on a general interval $[a, b]$ is constructed as
\begin{equation}\label{Gauss_Jacobi_2}
\int_a^b (b-x)^{\beta_{1}}(x-a)^{\beta_{2}} f(x)\ dx \approx \sum_{k=1}^{N} w_{k}^{(\beta_{1},\beta_{2})} f(x_{k}^{(\beta_{1},\beta_{2})}),
\end{equation}
where the nodes and weights are obtained through affine transformation:
\begin{equation*}
x_{k}^{(\beta_{1},\beta_{2})} = \frac{b-a}{2} \widehat{x}_{k}^{(\beta_{1},\beta_{2})} + \frac{a+b}{2}, \quad
w_{k}^{(\beta_{1},\beta_{2})} = \left( \frac{b-a}{2} \right)^{\beta_{1}+\beta_{2}+1} \widehat{w}_{k}^{(\beta_{1},\beta_{2})}.
\end{equation*}
This quadrature rule will be repeatedly used for the singular radial integrals arising in the fractional Laplacian and for the weakly singular temporal integrals in the Caputo derivative. Moreover, in the BRFE strategy (\cref{sec:computation_fractional_laplacian}), it is also employed to evaluate the loss function when the residual exhibits boundary singularities.

\subsection{Spatiotemporally separable neural network}\label{sec:STSNN}
This subsection introduces the STSNN architecture, which constitutes the main methodological extension beyond QE-MC-fPINN in the time-dependent setting. Its approximation properties and the computational complexity of related integral evaluations are closely connected to the tensor neural network in \cite{wang2024tensor}.

\begin{figure}[!htbp]
\centering
\includegraphics[width=14cm]{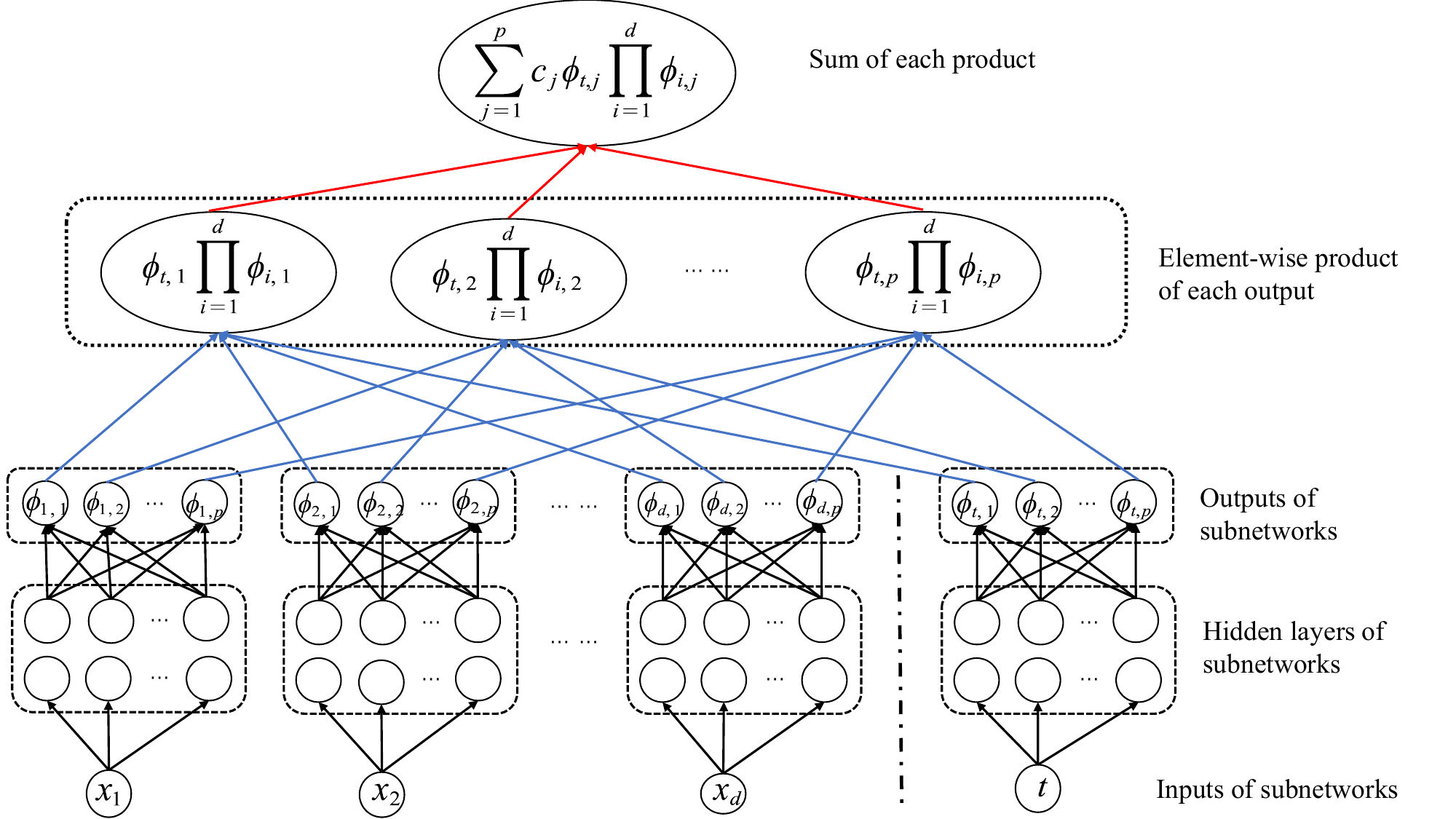}
\caption{Architecture of the spatiotemporally separable neural network (STSNN). Black arrows denote linear (or affine) transformations. Each blue arrow indicates that the ending node is the product of all starting nodes of the same color. The final output is obtained by summing the contributions from the red arrows.}\label{STSNN}
\end{figure}

The neural network is built with $d+1$ subnetworks, and each subnetwork is 
a continuous mapping from a bounded closed set $\Omega_{i}\subset\mathbb R$ ($i=1, \cdots, d$) to $\mathbb R^p$, which can be expressed as
\begin{align*}
\Phi_{i}(x_{i};\theta_{i})&=\big(\phi_{i,1}(x_{i};\theta_{i}), \phi_{i,2}(x_{i};\theta_{i}),\cdots,\phi_{i,p}(x_{i};\theta_{i})\big)^{\top},\\
\Phi_t(t;\theta_t)&=(\phi_{t,1}(t;\theta_t),\phi_{t,2}(t;\theta_t),\cdots ,\phi_{t,p}(t;\theta_t))^{\top}, 
\end{align*}
where each $x_{i}$ denotes the one-dimensional input, 
$\theta_{i}$ denotes the parameters of the $i$-th subnetwork, 
typically the weights and biases. As illustrated in  \cref{STSNN}, the neural network structure adopted in this paper is composed of $d$ fully connected neural networks (FNNs) 
for the spatial basis functions $\Phi_{i}(x_{i};\theta_{i}), i=1,2,\ldots,d$, 
and one FNN for the temporal basis function $\Phi_t(t;\theta_t)$. 

To improve numerical stability, we normalize each 
$\phi_{i,j}(x_{i})$, $\phi_{t,j}(t)$ and use the following normalized neural network structure:
\begin{equation}\label{def_NN_normed}
\Psi (\boldsymbol{x},t;c,\theta )=\sum_{j=1}^p{c_j}\widehat{\phi }_{t,j}(t;\theta_t)
\widehat{\varphi}_j(\boldsymbol{x};\theta_{\boldsymbol{x}}), 
\end{equation}
where 
\begin{equation*}
\widehat{\varphi}_j(\boldsymbol{x};\theta_{\boldsymbol{x}})
=\prod\limits_{i=1}^d{\widehat{\phi }_{i,j}(x_{i};\theta_{i})},\:
\widehat{\phi }_{i,j}(x_{i};\theta _{i})=\frac{\phi _{i,j}(x_{i};\theta _{i})}
{\left\| \phi _{i,j}(x_{i};\theta _{i}) \right\| _{L^{2}(\Omega _{i})}}
, \: \widehat{\phi }_{t,j}(t;\theta _t)=\frac{\phi _{t,j}(t;\theta _t)}
{\left\| \phi _{t,j}(t;\theta _t) \right\| _{L^{2}((0,T])}}.
\end{equation*} 
Here, the \(L^2\) norms are precomputed using high-order Gauss quadrature on each \(\Omega_i\) and on the time interval \((0,T]\).

We denote the neural network parameters as $\theta=\{\theta_t,\theta_{\boldsymbol{x}}\}=\{\theta_t,\theta_1,\theta_2,...,\theta_d\}.$
To simplify the notation, we drop the hats hereafter and write \(\phi_{i,j}\), \(\phi_{t,j}\), and \(\varphi_j\) for the normalized basis functions unless otherwise stated. 

Owing to its separable structure, the proposed network can be viewed as a CANDECOMP/PARAFAC decomposition in \(L^2(\Omega\times(0,T])\) \cite{wang2024tensor}. For the present paper, this structure is crucial because it allows the time-space residual to be factorized into lower-dimensional temporal and spatial integrals, so dense temporal quadrature becomes feasible without sacrificing the deterministic spatial treatment of the fractional Laplacian.

To demonstrate the effectiveness of solving fPDEs using the proposed neural network method, we introduce the following approximation result for functions in the space $H^{m}(\Omega \times (0,T])$.
\begin{theorem} \cite{wang2024tensor}
Assume that each $\Omega_{i}$ is a bounded closed interval in $\mathbb R$ 
for $i=1, \cdots, d$, $\Omega=\Omega_1\times\cdots\times\Omega_{d}$, 
and the function $f(\boldsymbol{x},t)\in H^m(\Omega \times (0,T])$ with a non-negative integer $m$. 
Then for any tolerance $\varepsilon>0$, 
there exists a positive integer $p$ and the corresponding neural network basis functions defined by \cref{def_NN_normed} such that the following approximation property holds
\begin{equation*}
\|f(\boldsymbol{x},t)-\Psi(\boldsymbol{x},t;c,\theta)\|_{H^m(\Omega \times (0,T])}<\varepsilon.
\end{equation*}
\end{theorem}
This result guarantees the approximation of smooth functions. Since the possible boundary singularities are handled by the explicit boundary feature factor in the trial functions, the remaining smoother residual is then well approximated by the neural network components.

\subsection{Computation of the fractional Laplacian}\label{sec:computation_fractional_laplacian}
This subsection describes the spatial operator discretization. Compared with the QE-MC-fPINN method, this paper uses the same geometry-adaptive splitting, while the main difference lies in that the angular Monte Carlo sampling is replaced by deterministic directional quadrature. For simplicity of notation, we write $\varphi_j(\boldsymbol{x})$ for $\varphi_j(\boldsymbol{x};\theta_{\boldsymbol{x}})$ and $\phi_{t,j}(t)$ for $\phi_{t,j}(t;\theta_t)$.

\subsubsection{Discrete scheme of the fractional Laplacian}\label{sec:our_dis_sche}
We adopt a geometry-adaptive strategy proposed in \cite{ma2026quadrature} that splits the fractional Laplacian based on a spatially varying radius and directional distance-to-boundary information. For any \(\boldsymbol{x}\in\Omega\), define \(r_{0}(\boldsymbol{x})\) as the minimum distance from \(\boldsymbol{x}\) to the boundary \(\partial\Omega\). For each spatial basis function \(\varphi_j\), the fractional Laplacian is first divided into near-field and far-field parts:
\begin{equation*}
\frac{(-\Delta)^{\alpha/2} \varphi_j(\boldsymbol{x})}{C_{d,\alpha}} 
= \left(\int_{\|\boldsymbol{y}-\boldsymbol{x}\|_2 < r_{0}(\boldsymbol{x})} +\int_{\|\boldsymbol{y}-\boldsymbol{x}\|_2 \ge r_{0}(\boldsymbol{x})}\right) 
\frac{\varphi_j(\boldsymbol{x}) - \varphi_j(\boldsymbol{y})}{\|\boldsymbol{x} 
- \boldsymbol{y}\|_2^{d+\alpha}}\,d\boldsymbol{y}
:= I_{1,j}(\boldsymbol{x})+I_{2,j}(\boldsymbol{x}).
\end{equation*}
It is derived in \cite{ma2026quadrature} that the near-field integral
\begin{align*}
I_{1,j}(\boldsymbol{x})
&= r_{0}(\boldsymbol{x})^{-\alpha} \int_{\mathcal{S}_+^{d-1}} \int_0^1 \tau^{1-\alpha} \frac{F_{j}\left(\boldsymbol{x}, r_{0}(\boldsymbol{x})\tau, \boldsymbol{\xi}\right)}{\tau^2} \, d\tau \, J_{d}(\boldsymbol{\xi}) d\boldsymbol{\xi},
\end{align*}
with
\begin{equation*}
F_{j}(\boldsymbol{x}, r, \boldsymbol{\xi}) := 2\varphi_j(\boldsymbol{x}) - \varphi_j(\boldsymbol{x}+r\boldsymbol{\xi}) - \varphi_j(\boldsymbol{x}-r\boldsymbol{\xi}).
\end{equation*}
Here $\mathcal{S}_+^{d-1}$ denotes the upper hemisphere in $\mathbb{R}^d$.
To handle the singular behavior of $I_{1,j}(\boldsymbol{x})$ around \(\boldsymbol{y}=\boldsymbol{x}\), the Gauss-Jacobi quadrature is used:
\begin{align*}
I_{1,j}(\boldsymbol{x})\approx r_{0}(\boldsymbol{x})^{-\alpha}   \sum_{\ell=1}^{N_0} \sum_{k=1}^{N} w_\ell J_{d}(\boldsymbol{\xi}_\ell)  w_k^{(0,1-\alpha)} \frac{F_{j}
\left(\boldsymbol{x}, r_{0}(\boldsymbol{x})\tau_k^{(0,1-\alpha)}, \boldsymbol{\xi}_\ell\right)}{\left(\tau_k^{(0,1-\alpha)}\right)^{2}},
\end{align*}
where \(\{\tau_k^{(0,1-\alpha)}\}_{k=1}^{N}\) and \(\{w_k^{(0,1-\alpha)}\}_{k=1}^{N}\) are  Gauss-Jacobi points and weights on \([0,1]\). For \(d=2\), the angular integral over \(\mathcal{S}_+^{1}\) is discretized by an \(N_0\)-point Gauss-Legendre rule on \((0,\pi]\), producing the directional nodes \(\{\boldsymbol{\xi}_\ell\}_{\ell=1}^{N_0}\) and weights \(\{w_\ell\}_{\ell=1}^{N_0}\). For \(d=3\), a tensor-product Gauss-Legendre rule in the two spherical angles is used to generate the corresponding nodes and weights.

\begin{figure}[!tbhp]
\centering
\includegraphics[width=12cm]{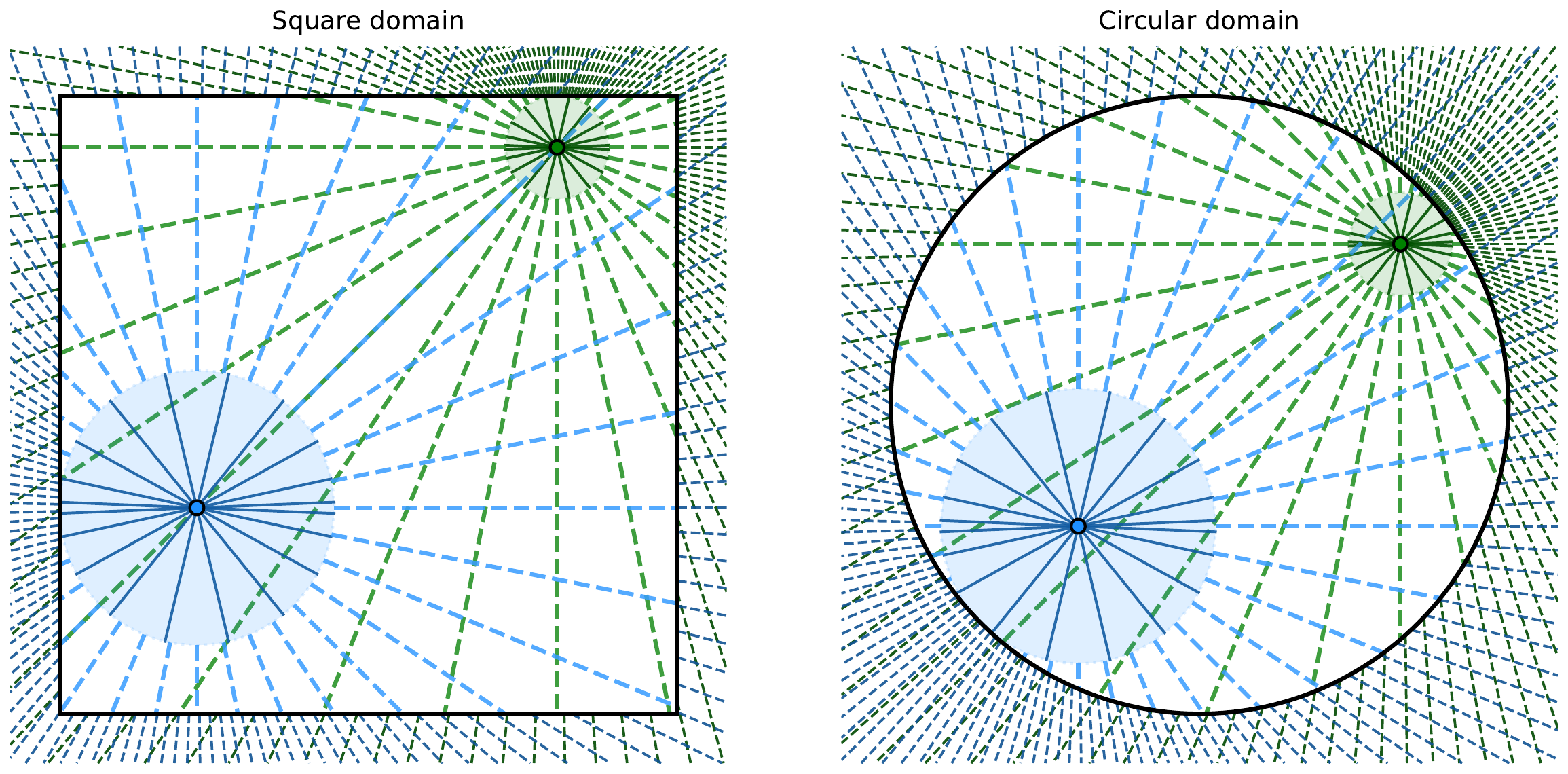}
\caption{Discretization of the fractional Laplacian on 2D domains.  
The black curve denotes the boundary \(\partial\Omega\). For each interior point \(\boldsymbol{x}\) (colored dots), a local ball \(B_{r_0(\boldsymbol{x})}(\boldsymbol{x})\) is defined. \(\mathbb{R}^2\) is split into three regions: (1) near-field from \(\boldsymbol{x}\) to \(\partial B_{r_0(\boldsymbol{x})}(\boldsymbol{x})\), using \(N_0=10\) symmetric Gaussian directions; (2) interior far-field from \(\partial B_{r_0(\boldsymbol{x})}(\boldsymbol{x})\) to \(\partial\Omega\), with \(N_1=32\) uniform directions; and (3) exterior far-field from \(\partial\Omega\) to infinity, with \(N_2=125\) uniform directions. Line segments show integration paths, and colors distinguish different evaluation points.
}\label{fig:2d_fra_dis}
\end{figure}
The distance from \(\boldsymbol{x}\) to the boundary along a unit direction \(\boldsymbol{\xi}\in\mathcal{S}^{d-1}\) was defined in \cite{ma2026quadrature} as
\begin{equation*}
d_{\boldsymbol{x}}(\boldsymbol{\xi}):=\min\{\,t>0 \mid \boldsymbol{x}+t\boldsymbol{\xi}\in\partial\Omega \,\}.
\end{equation*}
Then the far-field integral \(I_{2,j}\) is split into the part inside the domain and the part over the exterior domain. Using the quadrature rule in the radial direction of the inside part, we have
\begin{align*}
I_{2,j}(\boldsymbol{x})
&=\int_{\mathcal{S}^{d-1}} \left( \int_{r_{0}(\boldsymbol{x})}^{d_{\boldsymbol{x}}(\boldsymbol{\xi})} 
+ \int_{d_{\boldsymbol{x}}(\boldsymbol{\xi})}^{\infty} \right)
\frac{\varphi_j(\boldsymbol{x}) - \varphi_j(\boldsymbol{x}+r\boldsymbol{\xi})}{r^{1+\alpha}}\,dr\; J_{d}(\boldsymbol{\xi})d\boldsymbol{\xi}\\
&=\int_{\mathcal{S}^{d-1}} \left[\left(d_{\boldsymbol{x}}(\boldsymbol{\xi})-r_{0}(\boldsymbol{x})\right)\int_{0}^{1}
\frac{\varphi_j(\boldsymbol{x}) - \varphi_j\left(\boldsymbol{x}+r(\boldsymbol{x},\boldsymbol{\xi},t)\boldsymbol{\xi}\right)}
{r(\boldsymbol{x},\boldsymbol{\xi},t)^{1+\alpha}}dt  + \frac{\varphi_j(\boldsymbol{x})}{\alpha\,[d_{\boldsymbol{x}}(\boldsymbol{\xi})]^{\alpha}} \right] J_{d}(\boldsymbol{\xi})d\boldsymbol{\xi}\\
&\approx\int_{\mathcal{S}^{d-1}} \left[\left(d_{\boldsymbol{x}}(\boldsymbol{\xi})-r_{0}(\boldsymbol{x})\right)\sum_{m=1}^{N'} w_{m}\frac{\varphi_j(\boldsymbol{x}) - \varphi_j\left(\boldsymbol{x}+r(\boldsymbol{x},\boldsymbol{\xi},t_{m})\boldsymbol{\xi}\right)}
	{r(\boldsymbol{x},\boldsymbol{\xi},t_{m})^{1+\alpha}}  + \frac{\varphi_j(\boldsymbol{x})}{\alpha\,[d_{\boldsymbol{x}}(\boldsymbol{\xi})]^{\alpha}} \right] J_{d}(\boldsymbol{\xi})d\boldsymbol{\xi}\\
&:=\int_{\mathcal{S}^{d-1}} \left[Q_{1,j}(\boldsymbol{x},\boldsymbol{\xi})+Q_{2,j}(\boldsymbol{x},\boldsymbol{\xi})\right] J_{d}(\boldsymbol{\xi})d\boldsymbol{\xi},
\end{align*}
where \(\{t_{m}\}_{m=1}^{N'}\) and \(\{w_{m}\}_{m=1}^{N'}\) are the Gauss-Legendre nodes and weights on \([0,1]\). Here $\mathcal{S}^{d-1}$ denotes the unit sphere in $\mathbb{R}^d$. Then $I_{2,j}$ is approximated by a discrete average over directions as follows:
\begin{align*}
I_{2,j}(\boldsymbol{x})\approx  \sum_{i=1}^{2}{\sum_{k=1}^{N_{i}}{\left(\pi/n_{i} \right) ^{d-1}J_{d}(\boldsymbol{\xi }_{ik})Q_{i,j}(\boldsymbol{x},\boldsymbol{\xi }_{ik})}},
\end{align*}
with $N_{i} = 2n_{i}^{d-1}$ the number of distributed directions. For $d\ge 2$, these directions are parameterized over the spherical coordinate domain $[0,2\pi] \times [0,\pi]^{d-2}$, while the associated weight for each direction is given by \(\left(\pi/n_{i} \right)^{d-1}J_{d}(\boldsymbol{\xi }_{ik})\). Direction vectors \(\boldsymbol{\xi}_{ik}\) in different dimensional cases were defined in \cite{ma2026quadrature}. 

So the fractional Laplacian is ultimately decomposed into three contributions: a singular near-field, a regular interior far-field, and an analytical exterior far-field. A schematic illustration for a two-dimensional domain is provided in \cref{fig:2d_fra_dis}. The discretization scheme described here can, in principle, be generalized to higher dimensions. However, this work focuses on the high-accuracy neural network method for real-world physical problems with dimensions up to three. Readers interested in neural network methods for higher dimensional fPDEs are referred to MC-fPINN \cite{guo2022monte}, Improved MC-fPINN \cite{hu2024tackling}, and QE-MC-fPINN method \cite{ma2026quadrature}, etc.

Although the present method adopts the same geometry-adaptive decomposition of the fractional Laplacian as QE-MC-fPINN, the two approaches differ fundamentally in network architecture and discretization, boundary treatment, accuracy characteristics, and optimization strategy. These  differences will be expounded one by one in the subsequent text.

First, in terms of network architecture and discretization, QE-MC-fPINN adopts a standard fully-connected PINN and employs Monte Carlo sampling for both the directional discretization of the fractional Laplacian and the loss evaluation, whereas the present method employs a tensor neural network that approximates a multivariate function as sums of products of univariate subnetworks, thereby enabling fully deterministic quadrature framework for both the angular discretization and the loss computation. Thus, QE-MC-fPINN is inherently stochastic and the present method is fully deterministic. For time-space fractional PDEs, the present method further employs a STSNN that factorizes the residual into independent temporal and spatial components, a structural capability unavailable in conventional PINNs.

Second, for boundary treatment, both methods adopt the boundary feature enhanced (BFE) strategy ($\mu_1 = \alpha/2$) to capture the operator-induced singularity. The present method further proposes the boundary and right-hand-side feature enhanced (BRFE) strategy: when the right-hand side (RHS) function is also singular, BRFE adjusts the exponent to $\mu_1 = \alpha + s$ and employs a weighted Gauss-Jacobi quadrature, thereby handling the joint singularity mechanism with more stable convergence and higher accuracy.

Third, with respect to dimensional applicability and accuracy, QE-MC-fPINN leverages the dimension-favorable scaling of Monte Carlo sampling and is well-suited to $d \ge 4$, while the present method trades higher per-dimension cost via deterministic directional quadrature for substantially superior accuracy, achieving relative $L^2$ errors one to three orders of magnitude lower on $d = 1, 2, 3$ benchmarks.

Fourth, regarding optimization, QE-MC-fPINN optimizes all parameters jointly via gradient descent, whereas the present method decouples the optimization, that is, linear coefficients are solved by least squares and only network weights are updated by gradient descent-improving conditioning and accelerating convergence.

In essence, the present method is a fully deterministic, high-accuracy solver tailored for low- to moderate-dimensional fractional PDEs, with adaptive boundary singularity treatment and subspace-based optimization that jointly distinguish it from the stochastic, Monte Carlo-based QE-MC-fPINN paradigm.

\subsubsection{Analysis of the discretization scheme}\label{sec:FL_Analysis}
In this subsection, we provide a detailed analysis of the discretization of fractional Laplacian to demonstrate the robustness and accuracy of the proposed method.

\paragraph{The near-field integral:}  
As derived in \cite{ma2026quadrature},
\begin{align*}
\int_0^{r_0(\boldsymbol{x})}\frac{F_j(\boldsymbol{x},r,\boldsymbol{\xi})}{r^{1+\alpha}}\,dr
=-\,\frac{r_0(\boldsymbol{x})^{2-\alpha}}{2-\alpha}\,\boldsymbol{\xi}^\top H(\varphi_j)(\boldsymbol{x})\boldsymbol{\xi}
+O(r_0(\boldsymbol{x})^{4-\alpha}), %&= -r_0(\boldsymbol{x})^{2-\alpha} \int_{0}^{1}\tau^{1-\alpha}\,\boldsymbol{\xi}^{\top}H(\psi_j)\boldsymbol{\xi} d\tau+ O(r_0(\boldsymbol{x})^{4-\alpha})\\ 
\end{align*}
justifying the use of Gauss-Jacobi quadrature with parameters \((0,1-\alpha)\), which is tailored for integrands with this type of weak singularity. Since \(B_{r_0(\boldsymbol{x})}(\boldsymbol{x})\subset\Omega\), the local Taylor expansion is valid for \(0<r\le r_0(\boldsymbol{x})\) under standard regularity assumptions. Combined with a deterministic angular quadrature, the discretization yields high-accuracy and computationally efficient evaluation of $I_{1,j}(\boldsymbol{x})$.

\paragraph{The far-field integral:}  
The angular integration over the unit sphere $\mathcal{S}^{d-1}$ is discretized using a uniform directional grid determined by the angular resolution parameter $n_{1}$, while the radial integration is treated separately for the terms $Q_{1,j}$ and $Q_{2,j}$. The term $Q_{2,j}$ admits an analytical expression, which allows a relatively dense directional sampling to be used without additional computational cost. In contrast, $Q_{1,j}$ is approximated by Gauss quadrature in the radial direction, and hence the principal error of $I_{2,j}$ arises from the discretization of the spherical integral of $Q_{1,j}$. For smooth integrands the uniform angular discretization yields an error decaying as $O(n_1^{-2})$ (see, e.g., \cite{Brauchart2014QMC,Grabner2018SphereError}), so the accuracy of $I_{2,j}$ can be improved by increasing $n_1$.

\subsection{Trial spaces and loss function for fractional Laplacian}
This subsection addresses the second bottleneck left open after improving operator discretization, namely the construction of trial spaces that remain accurate for boundary-singular solutions. For fractional Laplacian problems defined on bounded domains, a generic PINN ansatz is often inefficient because the residual loss and the boundary behavior are strongly coupled. As demonstrated in \cite{Grubb2013FractionalLO,Guo2023ADL}, solutions of fPDEs often exhibit an $\alpha/2$-order boundary singularity characterized by the asymptotic relation
\begin{equation*}
u(\boldsymbol{x}) \approx \operatorname{dist}(\boldsymbol{x},\partial\Omega)^{\alpha/2}u_{\mathrm{reg}}(\boldsymbol{x}),
\quad \boldsymbol{x}\in\overline{\Omega},
\end{equation*}
where \(u_{\mathrm{reg}}\) is comparatively smoother.
Accurate treatment of this boundary behavior is therefore indispensable for the fractional problems considered here \cite{Guo2023ADL,ma2026quadrature}.

To embed the homogeneous Dirichlet boundary condition directly into the ansatz, we introduce
a prescribed boundary feature function $b(\boldsymbol{x})$ satisfying
\begin{equation}\label{bd_fun}
b(\boldsymbol{x})>0,\:\boldsymbol{x}\in\Omega,\quad 
\mbox{and} \quad
b(\boldsymbol{x})=0, \:\boldsymbol{x}\in\Omega^{c},
\end{equation}
whose role is not only to impose the boundary condition, but also to provide an explicit mechanism for encoding the dominant boundary singularity inside the trial space.
The trial function \(\Psi(\boldsymbol{x}; c, \theta_{\boldsymbol{x}})\) which satisfies the homogeneous boundary conditions is then constructed as
\begin{equation}\label{eq:trai_fun}
\Psi (\boldsymbol{x};c,\theta_{\boldsymbol{x}} )
=\sum_{j=1}^p{c_j\,b(\boldsymbol{x})^{\mu_{j}}\prod_{i=1}^{d}\phi _{i,j}(x_{i};\theta _{i})}
:=\sum_{j=1}^p{c_j\,\varphi _j(\boldsymbol{x};\theta_{\boldsymbol{x}} ),}
\end{equation}
where $\{\mu_{j}\}_{j=1}^{p}$ denotes the exponents of the boundary feature functions. These exponents are the crucial parameters of the boundary-singularity-aware trial family. We use them to construct the approximation space according to the asymptotic boundary structure of the target solution.
To ensure the final output \(\Psi\) lies in the desired function space, 
we require that \(b(\boldsymbol{x})^{\mu_{j}}\in H_0^{\alpha/2}(\Omega)\).
The values of the sequence \(\{\mu_{j}\}_{j=1}^p\) are determined by specifying 
only its first and last elements, \(\mu_{1}\) and \(\mu_{p}\). 
The remaining parameters are then defined via linear interpolation:
\begin{equation}\label{eq:def_muj}
\mu_{j} = \mu_{1} + (j-1)(\mu_{p} - \mu_{1})/(p-1), \qquad j = 1, 2, \dots, p.
\end{equation}
The loss function is then defined as the \(L^{2}\) residual of the governing equation:
\begin{equation}\label{eq:loss_function}
\mathcal{L}(c,\theta_{\boldsymbol{x}}) 
= \left\| L_{\boldsymbol{x}}\Psi(\boldsymbol{x};c,\theta_{\boldsymbol{x}}) 
- f(\boldsymbol{x}) \right\|_{L^{2}(\Omega)}.
\end{equation}
Minimizing this loss with respect to the linear coefficients \(c\) and the network parameters \(\theta_{\boldsymbol{x}}\) yields the approximate solution. For problems involving fractional Laplacian, however, the integrand in \cref{eq:loss_function} may inherit a strongly singular boundary profile unless the dominant asymptotic factor is extracted explicitly. To this end, we define for \(\boldsymbol{x} \in \Omega\)
\begin{equation}\label{eq:s_condition}
f_{\mathrm{reg}}(\boldsymbol{x}) := f(\boldsymbol{x})/b(\boldsymbol{x})^{s}, \quad\mbox{with}
\quad\limsup_{\substack{\boldsymbol{x} \to \partial \Omega \\ \boldsymbol{x} \in \Omega}} \frac{f(\boldsymbol{x})}{b(\boldsymbol{x})^{s}} = C \neq 0.
\end{equation}
Here the \(\limsup\) is taken over all possible interior paths approaching the boundary. 
With this choice, \(f_{\mathrm{reg}}\) captures the remainder after the leading power-law boundary behavior has been removed. We assume \(s> -0.5\) which guarantees that the resulting residual is square-integrable and hence that the weighted least-squares formulation is numerically well posed.

The first exponent \(\mu_{1}\) in \cref{eq:def_muj} determines the dominant boundary behavior of the trial space, and therefore has a strong influence on the approximation quality. In this paper, we formulate a rule to align \(\mu_{1}\) with the prevailing singularity mechanism of the problem. When \(s \geq 0\), we employ the boundary feature enhanced (BFE) strategy which has been used in \cite{Guo2023ADL,ma2026quadrature}. When \(-0.5<s < 0\), the boundary and RHS feature enhanced (BRFE) strategy is used. That is
\begin{enumerate}
\renewcommand{\labelenumi}{(\alph{enumi})}
\item \textbf{BFE}: 
As in \cite{Guo2023ADL,ma2026quadrature}, we set \(\mu_{1}=\alpha/2\) to match the leading-order boundary profile of bounded-domain problems. This operator-driven choice applies when the RHS function introduces no additional boundary singularity. In this case, the loss is evaluated using standard Gauss-Legendre quadrature over \(\Omega\).

\item \textbf{BRFE}: We set $\mu_{1} = \alpha+s$. This choice incorporates the singularity index $s$ from the RHS function, aligning the trial space with the composite boundary asymptotics. This right-hand-side-aware selection applies when $s<0$. 
Then, the loss is approximated by a tensor-product Gauss-Jacobi rule as follows:
\begin{align*}
\mathcal{L}(c,\theta_{\boldsymbol{x}})^{2}
&=\left\|b(\boldsymbol{x})^{s}\left[b(\boldsymbol{x})^{-s}L_{\boldsymbol{x}}\Psi (\boldsymbol{x};c,\theta_{\boldsymbol{x}} )
-f_{\mathrm{reg}}(\boldsymbol{x}) \right] \right\|_{L^{2}(\Omega )}^{2}\\
&\approx\sum_{i_1=1}^{N_1}{\cdots 
\sum_{i_d=1}^{N_d}{\left( \prod_{k=1}^d{w_{i_k}^{(2s, 2s)}} \right) 
T^{2}\left( x_{1,i_1}^{(2s, 2s)},\dots ,x_{d,i_d}^{(2s, 2s)} \right)}}.
\end{align*}
\end{enumerate}
Here
\begin{equation*}
T(\boldsymbol{y})=\sum_{j=1}^p{c_j\left[b(\boldsymbol{y})^{-s}
L_{\boldsymbol{x}}\varphi_j(\boldsymbol{y};\theta_{\boldsymbol{x}} ) \right]}-f_{\mathrm{reg}}(\boldsymbol{y}),
\end{equation*}
where $\{x_{k,i_k}^{(2s, 2s)}\}$ and $\{w_{i_k}^{(2s, 2s)}\}$ are the
nodes and weights of the Gauss-Jacobi quadrature. 
This weighted formulation is especially important when \(s<0\), because the deterministic quadrature can resolve the singular residual accurately and makes the BRFE trial space numerically stable in the regime where a standard residual formulation becomes unreliable.

For non-homogeneous boundary conditions, 
a function $g(\boldsymbol{x})\in H^1(\Omega)$ 
satisfying
\begin{equation*}
g(\boldsymbol{x})=u_b(\boldsymbol{x} ), 
\ \ \boldsymbol{x}\in \partial\Omega,
\end{equation*}
is introduced. Then using the decomposition
$u(\boldsymbol{x}) = \widehat{u}(\boldsymbol{x}) +  g(\boldsymbol{x} )$, 
the non-homogeneous boundary value problem is transformed 
to a homogeneous one for  the solution $\widehat{u}(\boldsymbol{x}) $.

\subsection{Trial spaces and loss function for time-space fractional PDEs}
For the time-space fractional PDE, 
the trial function $\Psi$ in \cref{eq:trai_fun} is then redefined as
\begin{equation}\label{eq:Psi_time}
\Psi(\boldsymbol{x}, t; c, \theta) = \widehat{\Psi}(\boldsymbol{x}, t; c, \theta)+ u_0(\boldsymbol{x}),\quad \widehat{\Psi}(\boldsymbol{x}, t; c, \theta):=\sum_{j=1}^{p} c_j t^{\gamma} \phi_{t,j}(t) \varphi_j(\boldsymbol{x}).
\end{equation}
This design incorporates the factor \(t^{\gamma}\) into time basis functions $ \phi_{t,j}(t)$, 
following the approach proposed in \cite{lin2025TFPIDE}. Combined with the boundary-singularity-aware spatial features, this yields the time-space ansatz on which the STSNN subspace method is constructed. Substituting this decomposition into \cref{eq:fPDE} transforms the original problem into a homogeneous initial-boundary value problem for \(\widehat{\Psi}\) with RHS function \(\widehat{f}\).
The corresponding loss function is then given by
\begin{equation}\label{eq:time_space_loss}
\mathcal{L}(c,\theta )=\left\| \partial^\gamma \widehat{\Psi}(\boldsymbol{x},t;c,\theta)/\partial t^\gamma+L_ {\boldsymbol{x}}\widehat{\Psi}(\boldsymbol{x},t;c,\theta)-\widehat{f}(\boldsymbol{x},t) \right\|_{L^{2}(\Omega \times(0,T])}.
\end{equation}
The BFE and BRFE can be extended to this setting in the same manner as that for fPE. We omit the details for brevity.
For more information about the treatment for non-homogeneous initial/boundary value conditions, please refer to \cite{lin2025TFPIDE,wang2024solving}.

\section{fTNN: a tensor neural network subspace method for fPDEs}\label{Sec:Optimization}
Based on the neural network architecture, the discretization of the fractional Laplacian, and the associated loss constructions, we propose in this section a neural network subspace method for fPDEs. We first determine the dominant boundary singularity index \(s\) based on the RHS function. According to the sign of \(s\), we adopt either the BFE strategy (\(s\ge 0\)) or the BRFE strategy (\(s<0\)) to choose the parameter \(\mu_1\). The remaining parameters \(\mu_j\) (\(j=2,\dots,p-1\)) are then obtained recursively from \cref{eq:def_muj}, while \(\mu_p\) is prescribed as a fixed constant. Using these parameters, we construct a neural-network trial subspace spanned by the output and feature functions, and compute the approximate solution in this subspace in the least-squares sense. The subspace is iteratively updated during training to improve the approximation accuracy. 

We first present the STSNN subspace method for time-space fractional PDEs and then describe the corresponding neural network subspace method for the fPEs. Since we use tensor neural network to solve the fractional PDEs, we name this method fTNN.

\subsection{Time-space fractional PDEs}\label{sec:time_FPDE}
Substituting \cref{eq:Psi_time} into \cref{eq:fPDE}, the original problem is then transformed into a homogeneous initial-boundary value problem for \(\widehat{\Psi}\), with the corresponding RHS function \(\widehat{f}\).
To decompose the high-dimensional integrals appearing in the loss function into lower-dimensional temporal and spatial integrals, we express the time derivative of order \(\gamma\) of \(\widehat{\Psi}\), the operator \(L_{\boldsymbol{x}}\) applied to \(\widehat{\Psi}\), and \(\widehat{f}\) in separable forms as follows:
\begin{equation}\label{eq:separable}
\left\{
\begin{aligned}
\frac{\partial^{\gamma}\widehat{\Psi}(\boldsymbol{x},t)}{\partial t^{\gamma}}
&=\sum_{j=1}^{p}
c_j\frac{\partial^{\gamma}\left(t^{\gamma}\phi_{t,j}(t;\theta_t)\right)}{\partial t^{\gamma}}
\varphi_j(\boldsymbol{x};\theta_{\boldsymbol{x}})
:=
\sum_{j=1}^{p} c_j L_{t,j}^{1}(t)L_{\boldsymbol{x},j}^{1}(\boldsymbol{x}),\\
L_{\boldsymbol{x}}\widehat{\Psi}(\boldsymbol{x},t)
&=\sum_{j=1}^{p}
c_jt^{\gamma}\phi_{t,j}(t;\theta_t)\,
L_{\boldsymbol{x}}\varphi_j(\boldsymbol{x};\theta_{\boldsymbol{x}})
:=\sum_{j=1}^{p} c_j L_{t,j}^{2}(t)L_{\boldsymbol{x},j}^{2}(\boldsymbol{x}),\\
\widehat{f}(\boldsymbol{x},t)
&:=\sum_{r=1}^{p_1}
\widehat{f}_{t,r}(t)\widehat{f}_{\boldsymbol{x},r}(\boldsymbol{x}),
\end{aligned}
\right.
\end{equation}
where \(p_1\) denotes the number of terms in the spatiotemporally separable form of the source.

After that, we determine the boundary singularity index \(s\) from \cref{eq:s_condition}. 
If \(s\ge 0\), we adopt the BFE strategy, setting \(\mu_1=\alpha/2\) and using Gauss quadrature for the trial function and the loss function.  
If \(s<0\), we adopt the BRFE strategy, setting \(\mu_1=\alpha+s\) and using Gauss-Jacobi quadrature with parameters \((2s,2s)\).
After prescribing \(\mu_p\), we define the neural network subspace as
\begin{equation*}
V'_{p}(\theta)
:=\operatorname{span}\left\{
t^{\gamma}\phi_{t,j}(t;\theta_t)\,\varphi_j(\boldsymbol{x};\theta_{\boldsymbol{x}}),
\; j=1,\dots,p
\right\},\: \mbox{with} \: \varphi_j(\boldsymbol{x};\theta_{\boldsymbol{x}})=b(\boldsymbol{x})^{\mu_j}\prod_{i=1}^d \phi_{i,j}(x_i;\theta_i).
\end{equation*}

The approximate solution is then sought in the form \(u_p=\widehat{u}_p+u_0\) with \(\widehat{u}_p\in V'_p\), by minimizing the residual in the least-squares sense. More precisely, we seek \(\widehat{u}_p\in V'_p\) such that
\begin{equation}\label{eq:least_squares}
(L_{t,\boldsymbol{x}}\widehat{u}_p,L_{t,\boldsymbol{x}}\widehat{v}_p)=(\widehat{f},L_{t,\boldsymbol{x}}\widehat{v}_p),
\quad \forall\,\widehat{v}_p\in V'_p,
\end{equation}
where \((\cdot,\cdot)\) denotes the \(L^2\) inner product over the space-time domain \(\Omega\times(0,T]\).

After the \(\ell\)-th training step, the trial function  \(\Psi(\boldsymbol{x},t;c,\theta^{(\ell)})\) belongs to the subspace
\[
V'_{p}(\theta^{(\ell)})
:=
\mathrm{span}
\left\{
t^\gamma \cdot \phi_{t,j}(t;\theta_t^{(\ell)})\cdot\varphi_j(\boldsymbol{x};\theta_{\boldsymbol{x}}^{(\ell)}),
\; j=1,\dots,p
\right\},
\]
where \(\phi_{t,j}(t;\theta_t^{(\ell)})\) and \(\varphi_j(\boldsymbol{x};\theta_{\boldsymbol{x}}^{(\ell)})\) are defined by \cref{eq:Psi_time,eq:trai_fun}, respectively. Once the coefficients \(c\) are computed, the approximate solution \(\widehat{u}_p\) is obtained, and the subspace \(V'_p(\theta^{(\ell+1)})\) is updated through the loss optimization.

Assembling the discrete system on \(V'_p(\theta^{(\ell)})\) and exploiting the spatiotemporal separability of the basis and source representations yields the following stiffness matrix and load vector for the BFE (\(s\ge0\)) and BRFE (\(s<0\)) strategies:
\begin{align}
\label{eq:A_time}
A_{m,n}^{(\ell)}
&=\begin{cases}
\displaystyle
\sum_{j=1}^{2}\sum_{k=1}^{2}
\left(L_{t,n}^{k},L_{t,m}^{j}\right)_t\cdot
\left(L_{\boldsymbol{x},n}^{k},L_{\boldsymbol{x},m}^{j}\right)_{\boldsymbol{x}}, & s\ge 0,\\[1.2ex]
\displaystyle
\sum_{j=1}^{2}\sum_{k=1}^{2}
\left(L_{t,n}^{k},L_{t,m}^{j}\right)_t\cdot
\left(b^{-s}L_{\boldsymbol{x},n}^{k},b^{-s}L_{\boldsymbol{x},m}^{j}\right)_{\boldsymbol{x},
b^{2s}}, & s<0,
\end{cases}\\
\label{eq:B_time}
B_m^{(\ell)}
&=\begin{cases}
\displaystyle
\sum_{r=1}^{p_1}\sum_{j=1}^{2}
\left(\widehat{f}_{t,r},L_{t,m}^{j}\right)_t\cdot
\left(\widehat{f}_{\boldsymbol{x},r},L_{\boldsymbol{x},m}^{j}\right)_{\boldsymbol{x}}, & s\ge 0,\\[1.2ex]
\displaystyle
\sum_{r=1}^{p_1}\sum_{j=1}^{2}
\left(\widehat{f}_{t,r},L_{t,m}^{j}\right)_t\cdot
\left(b^{-s}\widehat{f}_{\boldsymbol{x},r},\,b^{-s}L_{\boldsymbol{x},m}^{j}\right)_{\boldsymbol{x},\,b^{2s}}, & s<0,
\end{cases}
\end{align}
for \(1\le m,n\le p\). Here, \((\cdot,\cdot)_t\) and \((\cdot,\cdot)_{\boldsymbol{x}}\) denote the temporal and spatial inner products, respectively, both approximated by Gauss quadrature. For the BRFE strategy, the boundary singular factor is absorbed into the spatial terms via the factor \(b^{-s}\), which naturally leads to the weighted spatial inner product
\begin{equation}\label{eq:inner_weight}
(u,v)_{\boldsymbol{x},\,b^{2s}}
:=\int_{\prod_{i=1}^{d}(a_i,b_i)}
u(\boldsymbol{x})\,v(\boldsymbol{x})\,b(\boldsymbol{x})^{2s}\,d\boldsymbol{x}.
\end{equation}
The weighted integral in \cref{eq:inner_weight} is evaluated by Gauss-Jacobi quadrature with parameters \((2s,2s)\).

The parameters \(\theta^{(\ell)}\) determine the trial subspace \({V'_p}^{(\ell)}\), while the coefficient vector \(c^{(\ell)}\) determines the approximation within that subspace. Consequently, the optimization procedure can be naturally divided into two substeps.

First, with the neural network parameters \(\theta^{(\ell)}\) fixed, the optimal coefficient vector \(c^{(\ell+1)}\) is obtained in the least-squares sense from \cref{eq:least_squares} by solving the linear system
\begin{equation}\label{linear_system_fpde}
	A^{(\ell)}c^{(\ell+1)}=B^{(\ell)}.
\end{equation}

Second, with the coefficient vector \(c^{(\ell+1)}\) fixed, the neural network parameters \(\theta^{(\ell+1)}\) are updated by minimizing the loss function  
\begin{equation}\label{eq:loss_ite}
\mathcal{L}^{(\ell+1)}(c^{(\ell+1)},\theta^{(\ell)})
=\left(c^{(\ell+1)}\right)^{\top}A^{(\ell)}c^{(\ell+1)}
-2\left(B^{(\ell)}\right)^{\top}c^{(\ell+1)}
+(\widehat{f},\widehat{f})
\end{equation}
using gradient-based optimization methods such as Adam or L-BFGS.
Thus, for fixed \(\theta^{(\ell)}\), minimizing the discrete loss is equivalent to solving the normal equations \cref{linear_system_fpde}.

\begin{algorithm}[!htbp]
\caption{fTNN method for the time-space fractional PDE}
\label{Algorithm_fPDE}
\begin{enumerate}
\item 	 \textbf{Transform} \cref{eq:fPDE} into a homogeneous problem for \(\widehat{\Psi}\) using \cref{eq:Psi_time}, with RHS \(\widehat{f}\).

\item \textbf{Express} \(\partial_t^\gamma \widehat{\Psi}\), \(L_{\boldsymbol{x}}\widehat{\Psi}\), and \(\widehat{f}\) in spatiotemporally separable form, see \cref{eq:separable}.

\item \textbf{Select the strategy.}
Determine the boundary singularity index \(s\) from  \cref{eq:s_condition}.
\begin{itemize}
\item If \(s\ge 0\) (BFE), set \(\mu_1=\alpha/2\) and use Gauss quadrature points.
\item If \(s<0\) (BRFE), set \(\mu_1=\alpha+s\) and use Gauss-Jacobi quadrature with \((2s,2s)\).
\end{itemize}

\item \textbf{Initialize.}
Construct the initial trial function \(\Psi(\boldsymbol{x},t;c^{(0)},\theta^{(0)})\) defined in \cref{eq:trai_fun} with the chosen \(\mu_1\) and \(\mu_p\). Set the maximum number of iterations \(M\) and initialize \(\ell=0\).

\item \textbf{Repeat until convergence or \(\ell=M\):}
\begin{enumerate}
\item Using \(\mu_1\) and \(\mu_p\) to construct the trial subspace 
\[
V'_p(\theta^{(\ell)})
:=\operatorname{span}
\left\{
t^\gamma \phi_{t,j}(t;\theta_t^{(\ell)})\,
\varphi_j(\boldsymbol{x};\theta_{\boldsymbol{x}}^{(\ell)}),
\; j=1,\dots,p
\right\}.
\]
\item Assemble the stiffness matrix \(A^{(\ell)}\) and load vector \(B^{(\ell)}\) as in \cref{eq:A_time} and \cref{eq:B_time}.

\item Solve the linear system $A^{(\ell)}c=B^{(\ell)}$
and set \(c^{(\ell+1)}=c\).

\item Update \(\theta^{(\ell+1)}\) by minimizing the loss function \cref{eq:loss_ite}, with \(c^{(\ell+1)}\) fixed.
\item Set \(\ell\leftarrow \ell+1\).
\end{enumerate}
\end{enumerate}
\end{algorithm}

The whole procedure is summarized in \cref{Algorithm_fPDE}. This algorithm can be interpreted as an alternating optimization procedure for the unknown parameters \(c\) and \(\theta\) in the trial function \cref{eq:trai_fun}: the optimal coefficients \(c\) are determined in the least-squares sense, while the neural-network parameters are updated through the loss minimization.

\subsection{Fractional Poisson equation}

\begin{algorithm}[!htbp]
\caption{fTNN method for the fractional Poisson equation}
\label{Algorithm_fPE}
\begin{enumerate}
\item \textbf{Select the strategy.}
Determine the boundary singularity index \(s\) from \cref{eq:s_condition}.
\begin{itemize}
\item If \(s\ge 0\) (BFE), set \(\mu_1=\alpha/2\) and use Gauss quadrature points.
\item If \(s<0\) (BRFE), set \(\mu_1=\alpha+s\) and use Gauss-Jacobi quadrature with \((2s,2s)\).
\end{itemize}

\item \textbf{Initialize.}
Construct the initial trial function \(\Psi(\boldsymbol{x};c^{(0)},\theta^{(0)})\) defined in \cref{eq:trai_fun} with the chosen \(\mu_1\) and \(\mu_p\). Set the maximum number of iterations \(M\) and initialize \(\ell=0\).

\item \textbf{Repeat until convergence or \(\ell=M\):}
\begin{enumerate}
\item Define the trial subspace \cref{eq:v_p} using \(\mu_1\) and \(\mu_p\)
\[
V_p^{(\ell)}
:=
\operatorname{span}
\left\{
\varphi_j(\boldsymbol{x};\theta_{\boldsymbol{x}}^{(\ell)}),
\; j=1,\dots,p
\right\}.
\]

\item Assemble the stiffness matrix \(A^{(\ell)}\) and load vector \(B^{(\ell)}\) using the BFE or BRFE strategy, see \cref{eq:A_fPE,eq:B_fPE}.

\item Solve the linear system $A^{(\ell)}c=B^{(\ell)}$
and set \(c^{(\ell+1)}=c\).

\item Update \(\theta_{\boldsymbol{x}}^{(\ell+1)}\) by minimizing the loss function \(\mathcal{L}^{(\ell+1)}(c^{(\ell+1)},\theta_{\boldsymbol{x}}^{(\ell)})\) with \(c^{(\ell+1)}\) fixed.

\item Set \(\ell\leftarrow \ell+1\).
\end{enumerate}
\end{enumerate}
\end{algorithm}

For the fPE defined in \cref{eq:fra_lapla_problem}, we introduce the neural-network trial subspace
\begin{equation}\label{eq:v_p}
V_p(\theta_{\boldsymbol{x}})
:=\operatorname{span}
\left\{
\varphi_j(\boldsymbol{x};\theta_{\boldsymbol{x}}),
\; j=1,\dots,p
\right\},
\end{equation}
where the approximate solution is found by minimizing the residual in the least-squares sense. More precisely, we seek \(u_p\in V_p(\theta_{\boldsymbol{x}})\) such that
\begin{equation}\label{eq:least_square_method}
\left((-\Delta)^{\alpha/2}u_p,\,(-\Delta)^{\alpha/2}v_p\right)_{\boldsymbol{x}}
=\left(f,\,(-\Delta)^{\alpha/2}v_p\right)_{\boldsymbol{x}},
\quad
\forall\, v_p\in V_p(\theta_{\boldsymbol{x}}).
\end{equation}
After the \(\ell\)-th training step, the neural network \(\Psi(\boldsymbol{x};c,\theta_{\boldsymbol{x}}^{(\ell)})\) belongs to the subspace
\[V_p^{(\ell)}
:=\operatorname{span}
\left\{\varphi_j(\boldsymbol{x};\theta_{\boldsymbol{x}}^{(\ell)}),
\; j=1,\dots,p
\right\}.
\]
By assembling the discrete system over \(V_p^{(\ell)}\), the stiffness matrix and load vector corresponding to the BFE and BRFE formulations can be written in the unified form
\begin{align}
\label{eq:A_fPE}
A_{m,n}^{(\ell)}
&=\begin{cases}
\left((-\Delta)^{\alpha/2}\varphi_n^{(\ell)}, (-\Delta)^{\alpha/2}\varphi_m^{(\ell)}\right)_{\boldsymbol{x}}, & s\ge 0,\\[1ex]
\left(b^{-s}(-\Delta)^{\alpha/2}(\varphi_n^{(\ell)}), b^{-s}(-\Delta)^{\alpha/2}(\varphi_m^{(\ell)})\right)_{\boldsymbol{x},\,b^{2s}}, & s<0,
\end{cases}\\
\label{eq:B_fPE}
B_m^{(\ell)}
&=\begin{cases}
\left(f,\,(-\Delta)^{\alpha/2}\varphi_m^{(\ell)}\right)_{\boldsymbol{x}}, & s\ge 0,\\[1ex]
\left(b^{-s}f,\,b^{-s}(-\Delta)^{\alpha/2}(\varphi_m^{(\ell)})\right)_{\boldsymbol{x},\,b^{2s}}, & s<0,
\end{cases}
\end{align}
for \(1\le m,n\le p\). Here, \((\cdot,\cdot)_{\boldsymbol{x},\,b^{2s}}\) is the weighted spatial inner product defined in \cref{eq:inner_weight}.

The parameters \(\theta_{\boldsymbol{x}}^{(\ell)}\) determine the trial subspace \(V_p^{(\ell)}\), while the coefficient vector \(c^{(\ell)}\) determines the approximation within this subspace. Accordingly, \cref{Algorithm_fPE} can be interpreted as an alternating optimization procedure for fPEs. Since its implementation is analogous to \cref{Algorithm_fPDE}, we omit the details here for brevity.

\section{Numerical examples}\label{Sec:Numerical}
This section check the performance of the proposed method on a sequence of one-, two-, and three-dimensional benchmarks for which the deterministic spatial quadrature is computationally meaningful. 
All the experiments are conducted on NVIDIA GeForce RTX 4090 D GPUs.

The following two types of errors between the approximate 
solution $\Psi (\boldsymbol{x},t;c,\theta)$ and the exact solution $u$  
are used to measure the convergence behavior and accuracy of the examples in this section: 
\begin{itemize}
\item Relative $L^{2}$ error
\begin{equation*}
e_{L^{2}}:=\frac{\|u-\Psi(\boldsymbol{x},t;c^{*},\theta^{*})\|_{L^{2}(\Omega\times(0,T])}}
{\|u\|_{L^{2}(\Omega\times(0,T])}}.
\end{equation*}
This norm is evaluated by high precision numerical quadrature.

\item Relative $L^{2}$ test error
\begin{equation*}
e_{\mathrm{test}}:=\frac{\sqrt{\sum_{k=1}^{K}\left(\Psi(\boldsymbol{x}^{k},t^{k};c,\theta)-u(\boldsymbol{x}^{k},t^{k})\right)^{2}}}
{\sqrt{\sum_{k=1}^{K}\left(u(\boldsymbol{x}^{k},t^{k})\right)^{2}}},
\end{equation*}
where the test points $\{(\boldsymbol{x}^{k},t^{k})\}$ are placed on a uniform grid of $300^{d+1}$ points over $\Omega\times(0,T]$ for 1D and 2D, and on a uniform $30^{4}$ grid for 3D.
\end{itemize}
For time-independent problems, the same definitions for the errors are used with the $L^{2}$ norm taken over $\Omega$ and test points $\{\boldsymbol{x}^{k}\}$ uniformly distributed in $\Omega$.

\begin{table}[!htbp]
\centering
\caption{Neural network architectures and quadrature point configurations 
for different dimensions using BFE (Gauss points) and BRFE (Gauss-Jacobi points)}
\label{tab:nn_quad_params}
\begin{tabular}{cccc}
\toprule
\multicolumn{1}{c}{\multirow{2}{*}{$d$}} & \multicolumn{1}{c}{\multirow{2}{*}{\textbf{Each FNN}}}& \multicolumn{2}{c}{\textbf{Num of quadrature points}} \\
\cmidrule(r){3-4}
& & \textbf{BFE} & \textbf{BRFE}  \\
\midrule
1 & $[1,50,50,50]$ & $128$ & $48$ \\
2 & $[1,50,50,10]$ & $[16,16]$ & $[16,16]$ \\
3 & $[1,10,10,5\,\,\,]$ & $[10,10,10]$ & $[10,10,10]$ \\
\bottomrule
\end{tabular}
\end{table}

In the following numerical experiments, all FNNs consist of two hidden layers with the Tanh activation function applied to each layer. To accurately compute the integral in the loss function \cref{eq:loss_function} over the domain \(\prod_{i=1}^{d}[a_{i},b_{i}]\), we employ the quadrature point configurations specified in \cref{tab:nn_quad_params}. 
For the rectangular domain, we define
\begin{equation*}
b(\boldsymbol{x}) = \max\left(\prod_{i=1}^d (x_i-a_i)(b_i-x_i), 0\right),
\end{equation*}
which vanishes on the entire boundary, satisfying \cref{bd_fun}. 
\begin{remark}
Near a face of the rectangular domain, e.g., \(x_1=a_1\), \(b(\boldsymbol{x})\sim (x_1-a_1)\,C(\boldsymbol{x}')\), where \(C(\boldsymbol{x}')\) is smooth and strictly positive on that face. Near edges or corners, however, \(b(\boldsymbol{x})\) decays faster than the minimal distance from boundary: near a corner where \(x_i=a_i\), $i=i_{1}, i_{2}, 
\cdots, i_{q}$, it holds that \(b(\boldsymbol{x})\sim\prod_{i=i_{1}}^{i_{q}}(x_i-a_i)\).
Consequently, on lower-dimensional boundary strata,
\[
b(\boldsymbol{x})^{\alpha/2}=o\left(\operatorname{dist}(\boldsymbol{x},
\partial\Omega)^{\alpha/2}\right).
\]
This property makes \(b(\boldsymbol{x})\) a natural factor in the trial functions: it faithfully reproduces the leading singular behavior of solutions near faces, while automatically providing higher-order decay near edges and corners, thereby preventing over-singularity on lower-dimensional boundary strata. The remaining singular or complex features of the solution are left to be learned by the neural network components in the trial space.
\end{remark}

In all experiments we set \(\mu_{p} = \mu_{1} + 0.5\), which was found to provide a sufficiently rich approximation space for the tested parameter ranges. The number of basis functions \(p\) is set to the output dimension of each subnetwork (e.g., 50 for 1D, 10 for 2D, 5 for 3D).
During training, the neural network (for the fPE) or the STSNN (for the fPDE) is optimized using Adam with learning rate $0.003$ for 1000 epochs, then fine-tuned by L-BFGS with initial step size $0.1$ for 200 iterations (stationary) or 400 iterations (time-dependent).

\subsection{Fractional Poisson equations}
In this subsection, the fTNN refers to \cref{Algorithm_fPE}.
\subsubsection{One dimensional case}
For the numerical discretization of the fractional Laplacian, we employ \(N = 10\) Gauss-Jacobi quadrature 
points for the near-field radial integral and \(N_0 = 100\) 
Gauss quadrature points for the far-field radial integral.   

We first consider the one-dimensional 
fPE on the interval \((0,1)\):
\begin{equation}\label{eq:1d_fra}
(-\Delta)^{\alpha/2} u(x) = f(x), \quad x \in (0,1),
\end{equation}
subject to homogeneous Dirichlet boundary conditions \(u(0) = u(1) = 0\).
The forcing terms with the fabricated smooth solutions are given respectively by
\begin{equation*}
f(x)=\begin{cases}
\left( \ell_{\alpha}(x,2)-2\ell_{\alpha}(x,3)+\ell_{\alpha}(x,4) \right)/(2\cos\left(\pi\alpha\right)),&		u_{\text{exact}}=x^{2}(1-x)^{2},\\
\left( \ell_{\alpha}(x,3)-3\ell_{\alpha}(x,4)+3\ell_{\alpha}(x,5)-\ell_{\alpha}(x,6) \right)/(2\cos\left(\pi\alpha\right)), &u_{\text{exact}}=x^3(1-x)^3,
\end{cases}
\end{equation*}
where 
\begin{equation*}
\ell_{\alpha}(x,n) = \frac{\Gamma(n+1)}{\Gamma(n+1-\alpha)} \left( x^{n-\alpha} + (1-x)^{n-\alpha} \right).
\end{equation*}
Since $s=2-\alpha>0$ and $s=3-\alpha>0$ for the two cases, we use the BFE strategy. The results in \cref{tab:1d_smooth} show fTNN provides a clear accuracy advantage over fPINN in this smooth setting. 

\begin{table}[!htbp]\small
\centering
\caption{Relative $L^{2}$ test error $e_{\mathrm{test}}$ for \cref{eq:1d_fra} with smooth solutions.}\label{tab:1d_smooth}
\begin{tabular}{ccccc} 
\toprule
& \multicolumn{2}{c}{ $u = x^2(1-x)^2$} & \multicolumn{2}{c}{ $u = x^3(1-x)^3$} \\ 
\cmidrule(r){1-1} \cmidrule(r){2-3} \cmidrule(l){4-5}
$\alpha$  &fPINN &  fTNN  &fPINN  & fTNN \\ 
\midrule
%0.4  & 1.76e-1  &   1.26e-5   & 3.22e-02 &   1.57e-6\\
1.5  & 3.79e-2   &  2.92e-7   & 1.67e-02  &  6.94e-7 \\
1.9  & 2.98e-3   &  5.70e-7  & 2.83e-04  &  1.38e-6 \\
\bottomrule
\end{tabular}
\end{table}

We then consider a compactly supported function that is the sum of two bump functions with different exponents:
\begin{equation}\label{eq:compact_support_sum}
u_{\text{exact}}(x) = (1 - x^{2})_+^{\beta_{1}} + (1 - x^{2})_+^{\beta_{2}}, \quad \text{for } x \in [-1,1],\quad \beta_{1} \leqslant \beta_{2}.
\end{equation}
The regularity of \( u_{\text{exact}}\) is determined by the smaller exponent \( \beta_{1} \), say
\begin{equation*}
u_{\text{exact}}\in\begin{cases}
C^{\beta_{1} - 1,1}(\mathbb{R}) ,&		\beta_{1} \in \mathbb{N},\\
C^{\lfloor \beta_{1} \rfloor, \beta_{1} - \lfloor \beta_{1} \rfloor}(\mathbb{R}), & \beta_{1} \notin \mathbb{N}.
\end{cases}
\end{equation*}
\begin{table}[!htbp]\small
\centering
\caption{Summary of parameter regimes and strategy selection. }
\label{tab:cases_summary}
\begin{tabular}{lllc}
\toprule
\textbf{Case} & \multicolumn{1}{c}{Conditions} & Strategy & $\mu_{1}$ \\
\midrule
Case I   & $\beta_{1}=\alpha/2$, $\beta_{2}=\alpha/2$ or $\beta_{2}\geq\alpha $ & BFE  & $\alpha/2$ \\
\cdashline{1-4}[1pt/1pt] 			
Case II  & $\beta_{1}=\alpha/2$, $ \alpha/2<\beta_{2}<\alpha$   & BRFE  & $\beta_{2}$ \\
\cdashline{1-4}[1pt/1pt] 	
Case III & $\beta_{1}<\alpha/2$ (any $\beta_{2}\ge\beta_{1}$)   & BRFE  & $\beta_{1}$ \\
\cdashline{1-4}[1pt/1pt] 	
Case IV  & $\alpha /2<\beta_{1}<\alpha$ (any $\beta_{2}\ge\beta_{1}$)& BRFE  & $\beta_{1}$ \\
\cdashline{1-4}[1pt/1pt] 	
Case V   & $\beta_{1}\ge \alpha$ (any $\beta_{2}\ge\beta_{1}$) & BFE   & $ \alpha /2$ \\
\bottomrule
\end{tabular}
\end{table}

For \(\beta_{i}\neq \alpha/2\), 
\[
(-\Delta)^{\alpha/2}(1-x^{2})_{+}^{\beta_{i}} \sim C_{i}\,(1-x^{2})^{\beta_{i}-\alpha}, \quad |x|\to 1^{-}.
\]
Since \(\beta_{1}\le\beta_{2}\), the smallest exponent dominates, yielding
\[
f(x) \sim \text{const}\cdot (1-x^{2})^{\beta_{1}-\alpha}, \quad \beta_{1}<\beta_{2},\;\beta_{1}\neq\alpha/2.
\]
In the special case \(\beta_{1}=\beta_{2}= \alpha/2\), we have
\[
(-\Delta)^{\alpha/2} (1-x^{2})_{+}^{\alpha/2}
= 2^{\alpha}\Gamma\left(\alpha/2+1/2\right)\,\Gamma\left(\alpha/2+1\right)/\sqrt{\pi},
\]
which is a non-zero constant, so \(f\) remains bounded and non-zero near the endpoints.
\begin{table}[!tbhp]\small
\centering
\caption{Errors of fTNN to solve \cref{eq:1d_fra} with exact solution \cref{eq:compact_support_sum}.}	\label{tab:1D_BFE_BRFE_class}
\begin{tabular}{cccccc}
\toprule
\multicolumn{3}{c}{BRFE ($s<0$)} & \multicolumn{3}{c}{BFE ($s\geqslant0$)} \\
\cmidrule(lr){1-3} \cmidrule(lr){4-6}
Case: $( \alpha ,\beta_{1},\beta_{2})$ & $e_{L^{2}}$ & $e_{\text{test}}$ & Case: $( \alpha ,\beta_{1},\beta_{2})$ & $e_{L^{2}}$ & $e_{\text{test}}$ \\
\midrule
II: (0.40, 0.20, 0.30) & 6.266e-5 & 6.595e-5 & I: (0.20, 0.10, 0.10) & 3.156e-5 & 3.188e-3 \\	
II: (1.60, 0.80, 1.20) & 3.492e-5 & 3.657e-5 & I: (1.80, 0.90, 0.90) & 8.615e-7 & 8.389e-7 \\		
II: (1.10, 0.55, 0.88) & 1.922e-5 & 1.902e-5 & I: (1.60, 0.80, 1.90) & 2.859e-6 & 2.852e-6 \\
II: (1.90, 0.95, 1.60) & 2.932e-4 & 7.300e-6 & I: (0.90, 0.45, 0.90) & 3.854e-5 & 3.854e-5 \\
III: (0.50, 0.05, 0.10)& 8.875e-5 & 3.074e-3 & V: (0.10, 0.10, 0.20) & 1.501e-5 & 2.087e-3 \\
III: (0.40, 0.08, 0.30)& 5.980e-5 & 3.122e-3 & V: (0.20, 0.20, 0.30) & 2.889e-5 & 4.785e-5 \\
III: (0.20, 0.08, 0.30)& 5.465e-5 & 5.391e-5 & V: (0.40, 0.41, 1.20) & 1.885e-5 & 1.910e-5 \\
III: (0.70, 0.30, 1.30)& 1.804e-5 & 5.391e-5 & V: (0.40, 1.20, 1.90) & 7.622e-7 & 4.270e-6 \\
IV: (1.30, 0.92, 1.30) & 6.899e-6 & 6.436e-6 & V: (1.10, 1.15, 1.61) & 1.006e-6 & 1.009e-6 \\
IV: (1.42, 1.20, 3.30) & 8.611e-7 & 8.577e-7 & V: (1.80, 1.90, 2.30) & 1.596e-7 & 1.623e-7 \\
IV: (1.90, 1.50, 1.80) & 2.135e-7 & 2.158e-7 & V: (1.70, 1.99, 2.90) & 4.110e-7 & 4.027e-7 \\
IV: (1.90, 1.80, 2.20) & 2.182e-7 & 2.180e-7 & V: (1.90, 2.30, 3.20) & 6.370e-7 & 6.348e-7 \\
\bottomrule
\end{tabular}
\end{table}
\begin{figure}[!htbp]
\centering
\subfloat[I: $( \alpha ,\beta_{1},\beta_{2})=(0.20,0.10,0.10)$,  BFE.]{\includegraphics[width=7.5cm]{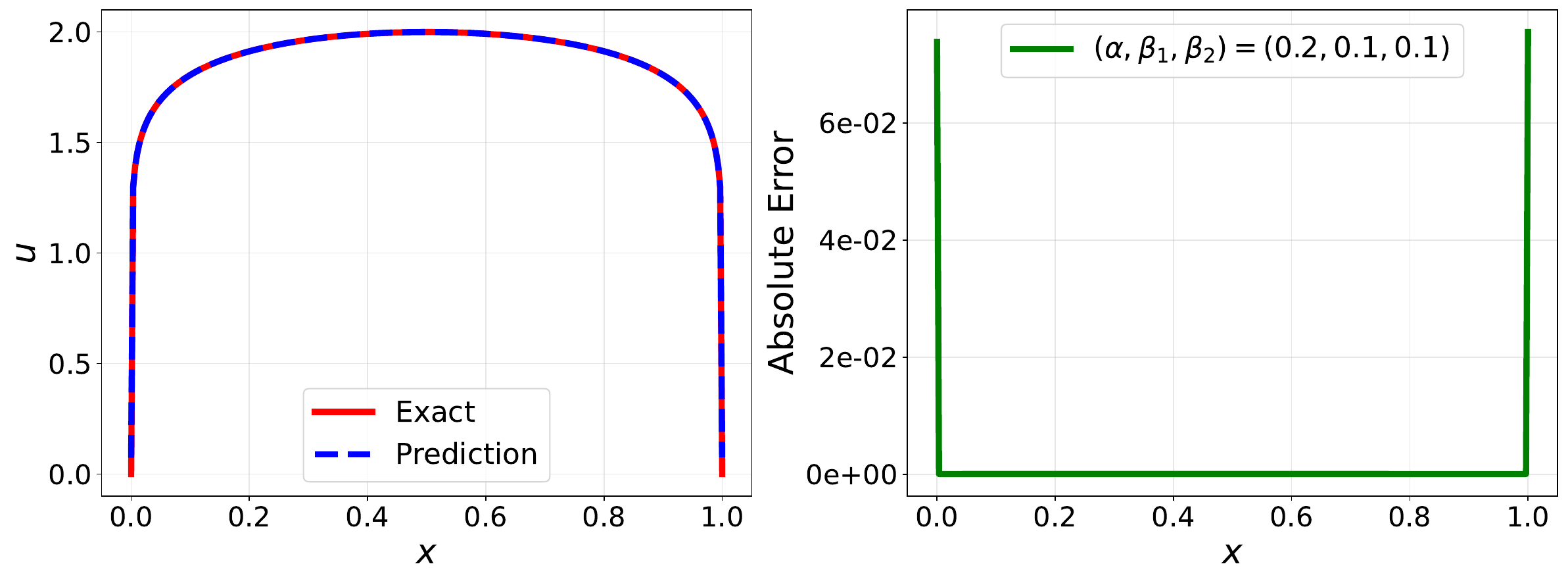}}
\hspace{.1cm}
\subfloat[II: $( \alpha ,\beta_{1},\beta_{2})=(1.90,0.95,1.60)$, BRFE.]{\includegraphics[width=7.5cm]{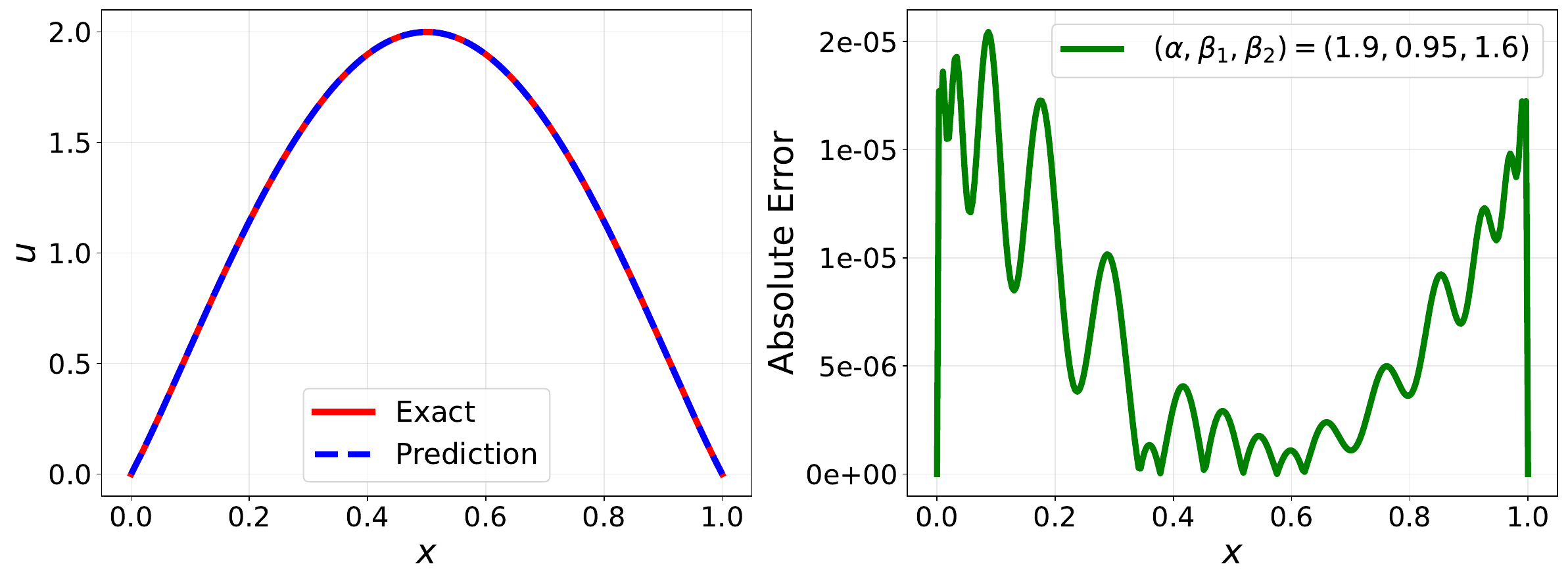}}\\
\subfloat[III: $( \alpha ,\beta_{1},\beta_{2})=(0.50,0.05,0.10)$, BRFE.]{\includegraphics[width=7.5cm]{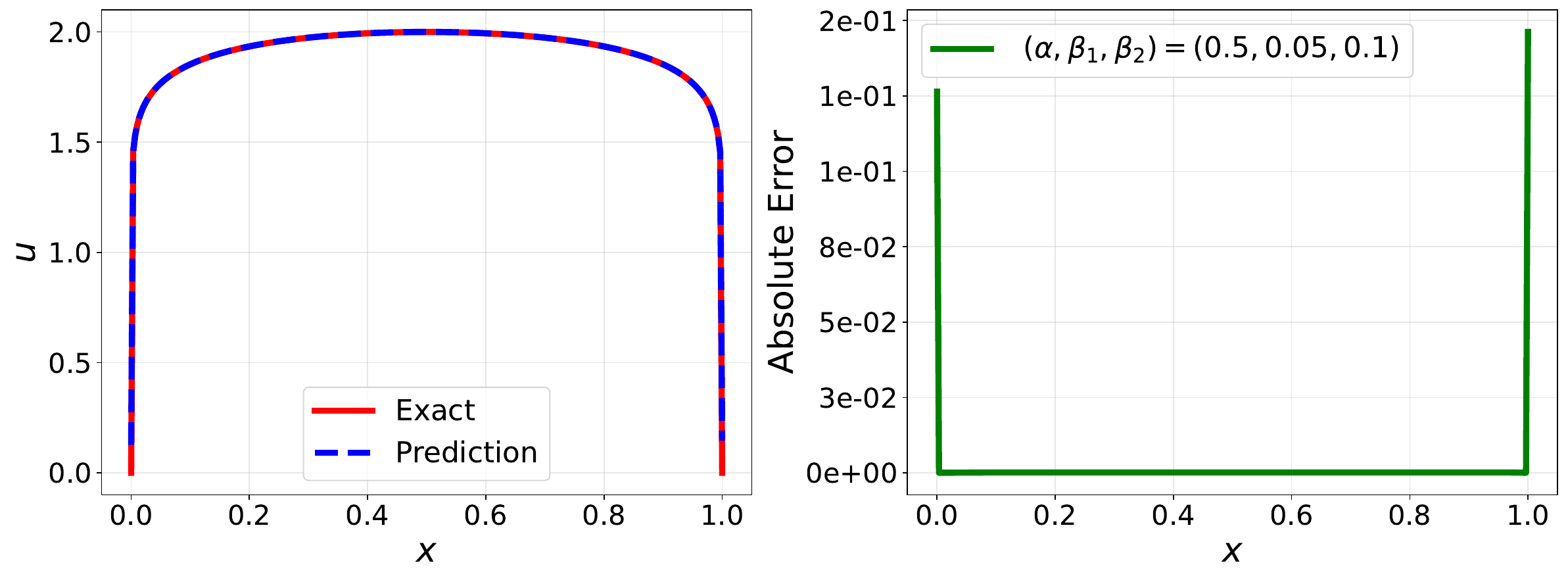}}\hspace{.1cm}
\subfloat[IV: $( \alpha ,\beta_{1},\beta_{2})=(1.90,1.80,2.20)$,  BRFE.]{\includegraphics[width=7.5cm]{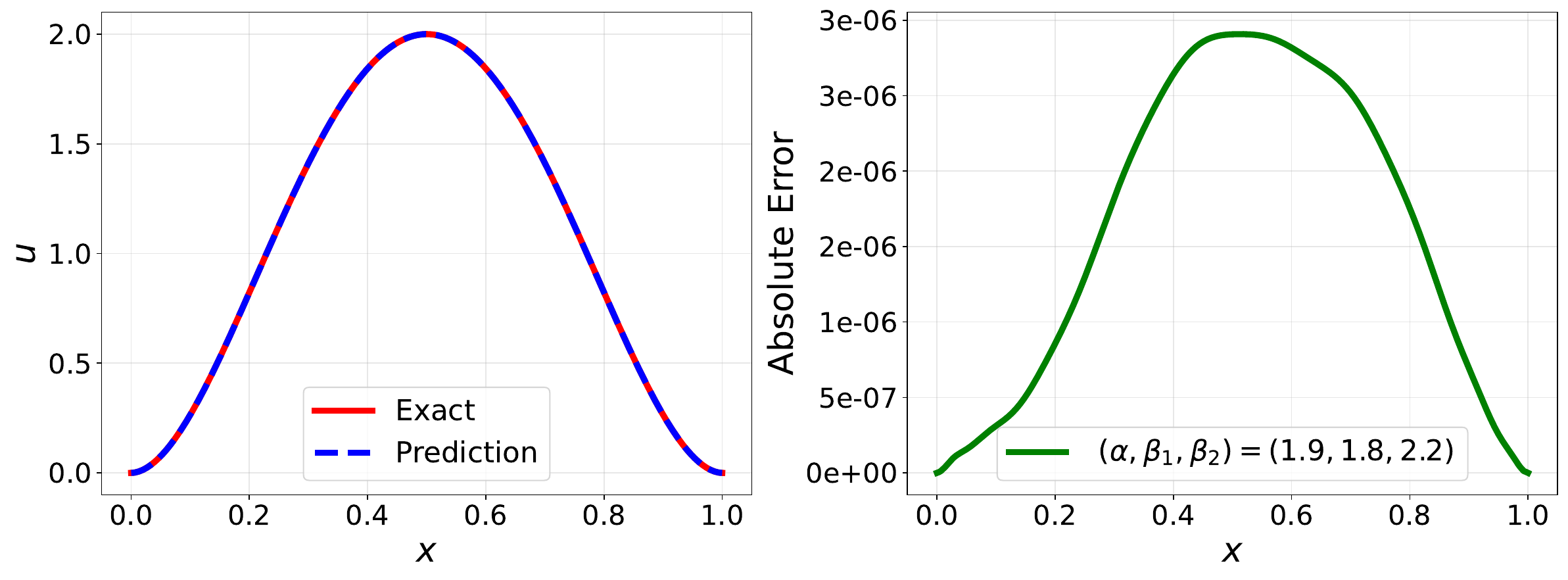}}\\
\subfloat[V: $( \alpha ,\beta_{1},\beta_{2})=(0.20,0.20,0.30)$, BFE.]{\includegraphics[width=7.5cm]{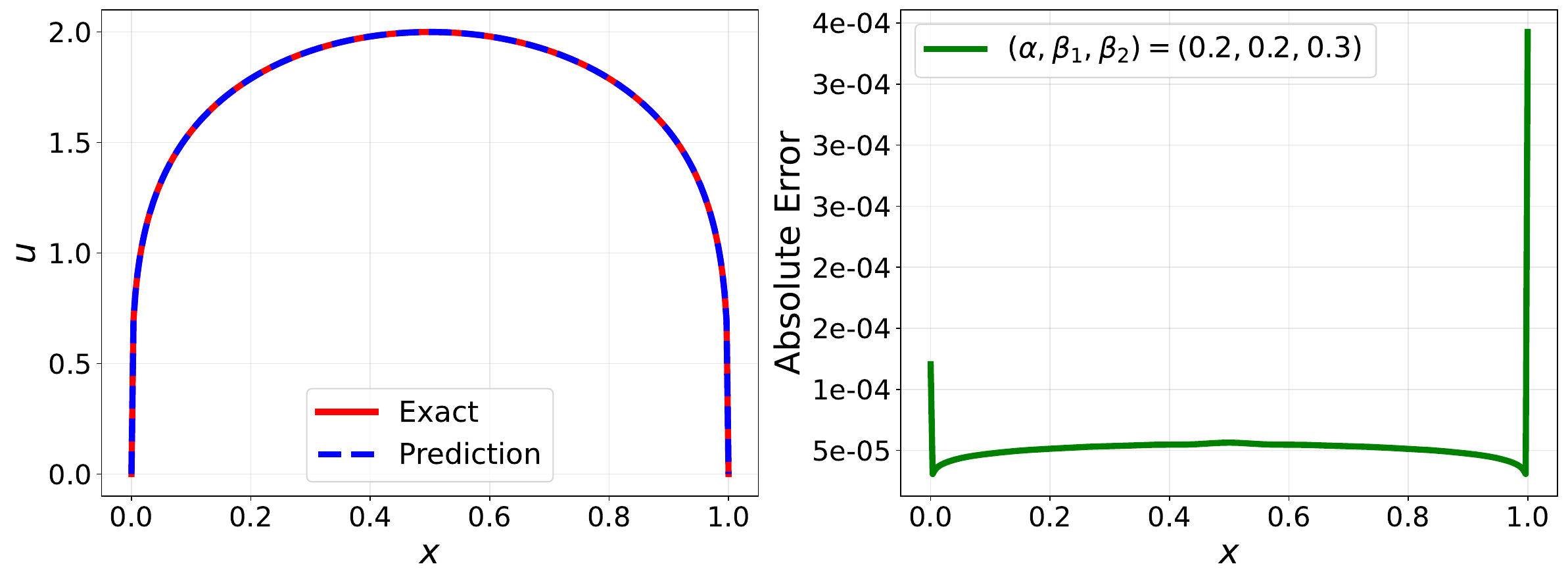}}\hspace{.1cm}
\subfloat[V: $( \alpha ,\beta_{1},\beta_{2})=(1.80,1.90,2.30)$,  BFE.]{\includegraphics[width=7.5cm]{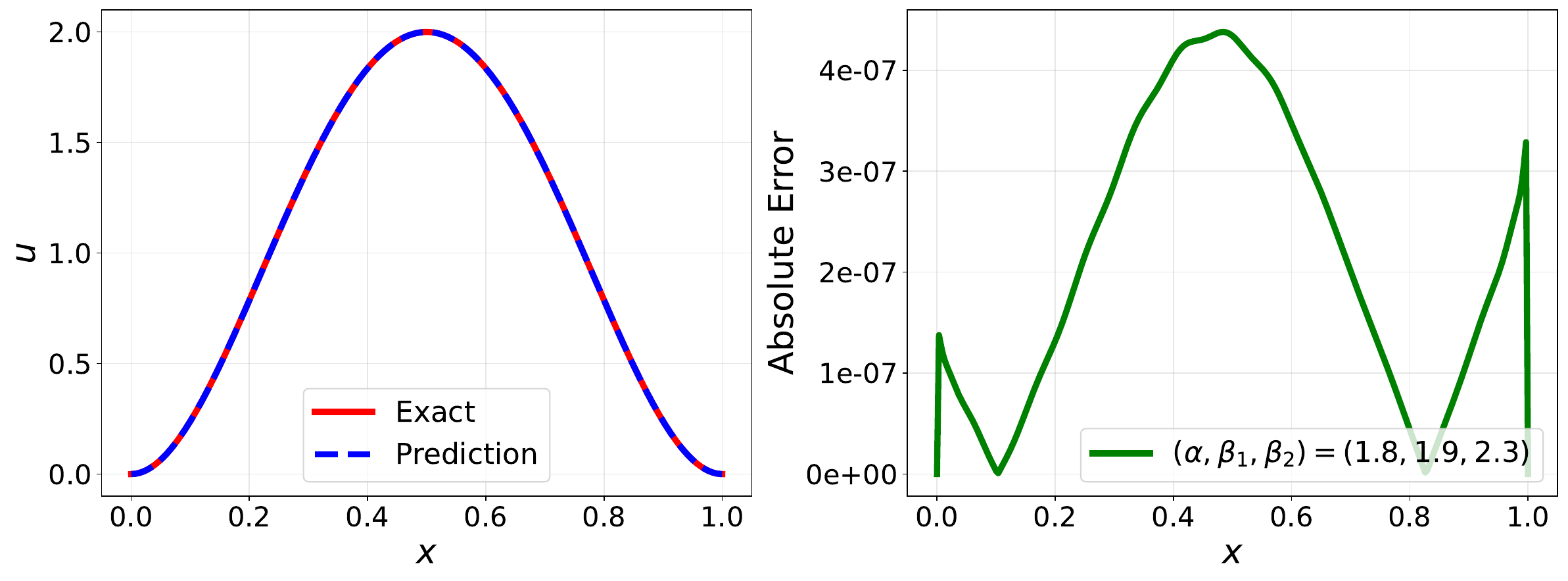}}
\caption{Numerical results of fTNN to solve \cref{eq:1d_fra} with exact solution \cref{eq:compact_support_sum}.}\label{fig:results}
\end{figure}

Consequently, the singularity index defined in \cref{eq:s_condition} is \(s=\beta_{1}-\alpha\), except in the special case \(\beta_{1}=\beta_{2}=\alpha/2\) where \(s=0\). The appropriate strategy, BFE for \(s\ge 0\) and BRFE for \(s<0\), then follows as summarized in \cref{tab:cases_summary}. 
Errors of fTNN are listed in \cref{tab:1D_BFE_BRFE_class} for five regimes with different parameter settings. As is shown, when \(s<0\), BRFE keeps the errors uniformly small by aligning the leading basis exponent with the composite singularity induced by the operator and the forcing term; when \(s\ge 0\), BFE recovers the canonical operator-driven boundary profile and achieves a comparable level of accuracy. In this sense, fTNN with BFE/BRFE achieves a further performance improvement compared with QE-MC-fPINN.
\begin{figure}[!htbp]
	\centering
	\includegraphics[width=12cm,height=4.2cm]{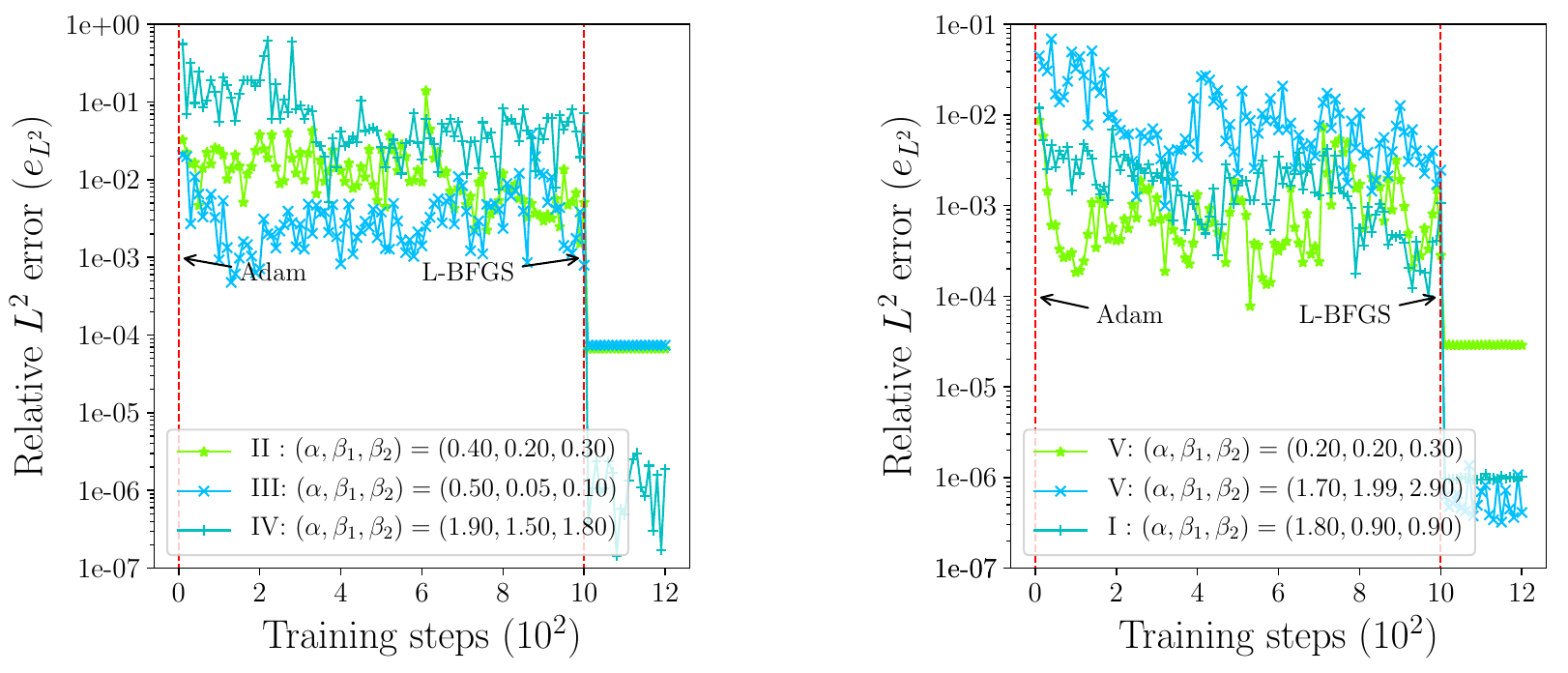}
	\caption{Relative $L^{2}$ errors of fTNN with BFE (left) and BRFE (right) strategies to solve \cref{eq:1d_fra} with exact solution \cref{eq:compact_support_sum}.}\label{fig:error_1d_curve}
\end{figure}

The solution profiles in \cref{fig:results} and the error histories in \cref{fig:error_1d_curve} show that fTNN with two strategies remains stable throughout training, with the largest errors concentrated near the endpoints where the singularity is the strongest. These results validate the robustness and accuracy of fTNN across the full range of fractional orders \(\alpha\) and exponent combinations \((\beta_{1},\beta_{2})\).
\subsubsection{FPEs on the unit square/cube}
We first consider the fPE \cref{eq:fra_lapla_problem} with an exact solution constructed as
\begin{equation}\label{eq:s_exact_23D}
u_{\text{exact}}(\boldsymbol{x})=\begin{cases}\quad 
\sum\limits_{k=1}^{2}{\prod\limits_{i=1}^{2}{\left(x_{i}-x_{i}^{2} \right) ^{\beta _k}}},
&\boldsymbol{x}\in [0,1]^{2},\\
100\sum\limits_{k=1}^{2}{\prod\limits_{i=1}^3{\left(x_{i}-x_{i}^{2} \right) ^{\beta _k}}},
&\boldsymbol{x}\in [0,1]^3.\\
\end{cases}
\end{equation}
where the exponents satisfy \( \beta_{1} \leqslant \beta_{2} \). 
Due to the complexity of deriving an explicit closed-form expression for the RHS function \(f(\boldsymbol{x}) = (-\Delta)^{\alpha / 2} u_{\text{exact}}(\boldsymbol{x})\), we resort to a numerical approximation to obtain a high-accuracy RHS function. To this end, we employ dense quadrature rules whose parameters are specified in \cref{tab:dis_para_23D}. 
\begin{table}[!htbp]\small
\centering
\caption{Parameters setting of fTNN to solve \cref{eq:fra_lapla_problem} with $f \approx (-\Delta)^{ \alpha /2} u_{\text{exact}}(\boldsymbol{x})$.}\label{tab:dis_para_23D}
\label{tab:dis_para_combined}
\begin{tabular}{cccccccc}
\toprule
\multicolumn{1}{c}{\multirow{2}{*}{\textbf{Integral}}} & \multicolumn{1}{c}{\multirow{2}{*}{\textbf{Parameters}}} & \multicolumn{1}{c}{\multirow{2}{*}{\textbf{Description}}} & \multicolumn{2}{c}{$(-\Delta)^{ \alpha /2} u_{\text{exact}}$} & \multicolumn{2}{c}{RHS \(f\)} \\ \cmidrule(lr){4-5} \cmidrule(lr){6-7}
\multicolumn{1}{c}{} & \multicolumn{1}{c}{} & \multicolumn{1}{c}{} & 2D & 3D & 2D & 3D \\
\midrule
\multirow{2}{*}{\(I_{1,j}(\boldsymbol{x})\)} 
& \(N_0\) & Angular directions on \(\mathcal{S}_{+}^{d-1}\) & 10 & \(120\) & 50 & \(300\) \\
& \(N\) & Gauss-Jacobi radial points & 10 & 8 & 200 & 32 \\
\midrule
\multirow{3}{*}{\(I_{2,j}(\boldsymbol{x})\)} 
& \(n_{1}\) & Angular resolution for \(Q_{1,j}\) & 32 & 16 & 600 & 35 \\
& \(n_2\) & Angular resolution for \(Q_{2,j}\) & 250 & 35 & 1000 & 35 \\
& \(N'\) & Gauss radial points for \(Q_{1,j}\) & 100 & 16 & 200 & 40 \\
\bottomrule
\end{tabular}
\end{table}

Although an explicit expression for the RHS function is unavailable, its singular behavior near the domain boundary can be characterized theoretically. This characterization is crucial for integration of the loss, as it enables the incorporation of the appropriate singular weight into the approximation space. By linearity, $f = f_{1} + f_{2}$ with 
\begin{equation*}
f_{k} = (-\Delta)^{ \alpha /2}\left(c_db(\boldsymbol{x})^{\beta_k}\right),\quad b(\boldsymbol{x}) = \prod\limits_{i=1}^d x_{i}(1-x_{i}),
\end{equation*}
on $\Omega = (0,1)^d$. Since $\beta_{1} \le \beta_{2}$, the singularity of $f$ is dominated by the term with the smaller exponent, i.e., $f \sim f_{1}$ as $\boldsymbol{x} \to \partial\Omega$. Near a face, say $x_{i} = 0$, we have $b(\boldsymbol{x}) \sim x_{i} B_{i}(\boldsymbol{x}')$, where $\boldsymbol{x}'$ denotes the remaining coordinates and $B_{i}(\boldsymbol{x}') = \prod_{j\neq i} x_{j}(1-x_{j})$ is smooth and non-zero on the face. Consequently, $b(\boldsymbol{x})^{\beta_{1}} \sim x_{i}^{\beta_{1}} \left(B_{i}(\boldsymbol{x}')\right)^{\beta_{1}}$. A classical result for the fractional Laplacian (see \cite{ros2016boundary}) yields the asymptotic expansion
\begin{equation*}
(-\Delta)^{ \alpha /2}\left(x_{i}^{\beta_{1}}\phi(\boldsymbol{x}')\right) \sim C(\beta_{1}, \alpha)\,\phi(\boldsymbol{x}')\,x_{i}^{\beta_{1}-\alpha}, \quad x_{i} \to 0^+,
\end{equation*}
where $C(\beta_{1},\alpha)$ is a constant. Applying this with $\phi = B_{i}^{\beta_{1}}$ gives
\begin{equation*}
f(\boldsymbol{x}) \sim C(\beta_{1},\alpha)\,B_{i}(\boldsymbol{x}')^{\beta_{1}}\,x_{i}^{\beta_{1}-\alpha},\quad \beta_{1} \le \beta_{2}.
\end{equation*}
Analogous expansions hold near other faces and edges, with the distance to the boundary raised to the appropriate power. Thus, near any face, $f(\boldsymbol{x})$ behaves like a constant (depending on the tangential coordinates) times $\operatorname{dist}(\boldsymbol{x},
\partial\Omega)^{\beta_{1}-\alpha}$. To extract this maximal singular behavior, we require that the limit defined in \cref{eq:s_condition} be finite and non-zero. Substituting the asymptotic forms shows that the exponent $s=\beta_{1}-\alpha$. With this choice, the ratio tends to a non-zero smooth function on the faces, while it vanishes near edges and corners, hence the $\limsup$ is indeed attained on the faces and equals a positive constant.

Based on the above asymptotic analysis, the selection of the strategy (BFE or BRFE) in the multi-dimensional setting follows a simpler criterion compared to the one-dimensional case. Unlike in 1D, where the special case \(\beta_{1}=\alpha /2\) yields a constant RHS function for the term \((1-x^{2})^{ \alpha /2}\), in higher dimensions the product structure \(\prod_{i} (x_{i}-x_{i}^{2})^{\beta_k}\) does not produce a constant fractional Laplacian for any \(\beta_k\).  This is because the cross-terms in the product lead to a genuine boundary singularity in $f$ even when $\beta_1 = \alpha/2$, justifying the exclusive use of $\beta_1 - \alpha$ as the exponent $s$ in the multidimensional setting. Consequently, the method choice is determined solely by the relative magnitudes of \(\beta_{1}\) and \(\alpha\):
\begin{itemize}
\item When \(\beta_{1} < \alpha\) (with any \(\beta_{2} \ge \beta_{1}\)), \(f(\boldsymbol{x})\) exhibits a genuine singularity \(\operatorname{dist}(\boldsymbol{x},
\partial\Omega)^{\beta_{1}-\alpha}\) near the boundary, necessitating the enriched approximation space of BRFE with the singular weight \(\mu_{1} = \beta_{1}\).
\item When \(\beta_{1} \ge \alpha\) (with any \(\beta_{2} \ge \beta_{1}\)), \(f(\boldsymbol{x})\) remains bounded (or even smooth) on \(\overline{\Omega}\), and the BFE with \(\mu_{1} =  \alpha /2\) suffices.
\end{itemize}

\begin{table}[!htbp]\small
\centering
\caption{Errors of fTNN to solve \cref{eq:fra_lapla_problem} on the unit square/cube.}\label{tab:combined_errors_fba_frba}
\begin{tabular}{cccccccc}
\toprule
\multicolumn{8}{c}{\textbf{2D: unit square}} \\
\multicolumn{4}{c}{BRFE ($s<0$), \(N_1=64\ (n_{1}=32)\) } & \multicolumn{4}{c}{BFE ($s \geqslant0$), \(N_1=64\ (n_{1}=32)\) }  \\
\cmidrule(lr){1-4} \cmidrule(lr){5-8}
$\alpha$ &  $(\beta_{1},\beta_{2})$ & $e_{L^{2}}$  &  $e_{\mathrm{test}}$ &  $\alpha$   & $(\beta_{1},\beta_{2})$ & $e_{L^{2}}$ & $e_{\mathrm{test}}$ \\
\midrule
0.50 & (0.08,0.20) & 3.030e-5 & 6.353e-3 & 0.20 & (0.20,0.40) & 3.497e-5 & 2.771e-4 \\
0.60 & (0.15,2.20) & 3.598e-5 & 1.237e-3 & 0.10 & (0.51,0.80) & 5.246e-5 & 5.306e-5 \\
1.00 & (0.55,0.80) & 6.233e-5 & 6.226e-5 & 0.80 & (0.90,1.20) & 7.308e-5 & 7.150e-5 \\
1.20 & (0.80,1.20) & 7.202e-5 & 7.203e-5 & 1.00 & (1.21,1.53) & 8.097e-5 & 7.850e-5 \\
1.60 & (1.50,1.79) & 3.160e-5 & 3.160e-5 & 1.20 & (1.80,1.90) & 7.726e-5 & 7.631e-5 \\
1.90 & (1.65,1.85) & 1.070e-5 & 1.070e-5 & 1.80 & (1.90,1.95) & 2.446e-5 & 2.362e-5 \\
1.96 & (1.45,2.30) & 4.088e-5 & 4.088e-5 & 1.76 & (1.70,2.20) & 3.470e-5 & 3.454e-5 \\
\midrule
\multicolumn{8}{c}{\textbf{3D: unit cube}} \\
\multicolumn{4}{c}{BRFE ($s<0$), \(N_1=512\ (n_{1}=16)\) } & \multicolumn{4}{c}{BFE ($s \geqslant0$), \(N_1=512\ (n_{1}=16)\) }  \\
\cmidrule(lr){1-4} \cmidrule(lr){5-8}
$\alpha$ &  $(\beta_{1},\beta_{2})$ & $e_{L^{2}}$  &  $e_{\mathrm{test}}$ &  $\alpha$   & $(\beta_{1},\beta_{2})$ & $e_{L^{2}}$ & $e_{\mathrm{test}}$ \\
\midrule
0.40 & (0.08,0.20) & 9.335e-4 & 2.936e-2 & 0.20 & (0.20,0.30) & 7.941e-4 & 1.118e-3 \\
1.00 & (0.55,0.80) & 4.838e-3 & 3.389e-3 & 0.80 & (0.80,1.80) & 6.025e-4 & 5.931e-4 \\
1.10 & (0.65,2.30) & 5.813e-3 & 4.128e-3 & 1.10 & (1.20,1.99) & 5.747e-4 & 5.689e-4 \\
1.20 & (0.90,1.70) & 9.087e-4 & 7.887e-4 & 1.20& (1.30,1.70) & 5.687e-4 & 5.646e-4 \\
1.60 & (1.20,1.75) & 3.369e-3 & 3.174e-3 & 1.40 & (1.50,1.85) & 4.874e-4 & 4.858e-4 \\
1.90 & (1.65,1.85) & 2.333e-4 & 2.306e-4 & 1.70 & (1.86,2.88) & 3.635e-4 & 3.637e-4 \\
1.96 & (1.65,2.45) & 1.495e-4 & 1.488e-4 & 1.90 & (1.90,2.80) & 1.005e-4 & 1.006e-4 \\
\bottomrule
\end{tabular}
\end{table}

The relative errors \(e_{L^{2}}\) and \(e_{\mathrm{test}}\) of fTNN to solve \cref{eq:fra_lapla_problem} on the unit square/cube are listed in \cref{tab:combined_errors_fba_frba}. Both types of errors lie below \(10^{-4}\) for most cases, confirming the high accuracy of fTNN in two and three dimensions. 

%For two representative scenarios, say \(\beta_{1} < \alpha\) (using BRFE strategy) and \(\beta_{1} \ge \alpha\) (using BFE strategy). 
%\Cref{fig:23d_error_curve} shows the decay of the relative \(L^{2}\) error during training, further demonstrating the reliability of the proposed method.
%\begin{figure}[!htbp]
%\centering
%\includegraphics[width=12cm]{fig/FBA_2d_0.2_0.3_s_0.1_theta12_64_10_N_j_10_50_mu_0.1_0.5.pdf}	
%\includegraphics[width=12cm]{fig/FRBA_2d_1.65_1.85_s_0.95_theta12_64_10_N_j_10_100_mu_1.65_0.1.pdf}
%\caption{Numerical results of fTNN to solve \cref{eq:fra_lapla_problem} on the unit square: $(\beta_{1},\beta_{2})=(0.20,0.30)$, $\alpha =0.20$ by BFE (top row);  $(\beta_{1},\beta_{2})=(1.65,1.85)$, $\alpha=1.90$ by BRFE (bottom row).}\label{fig:2d_fra_lap} 
%\end{figure} 
%
%\begin{figure}[!htbp]
%\centering
%\includegraphics[width=12cm,height=4.2cm]{fig/2d_square_error_loss.pdf}
%\includegraphics[width=12cm,height=4.2cm]{fig/3d_cube_error_loss.pdf}
%\caption{Losses and the relative $L^{2}$ errors of fTNN to solve \cref{eq:fra_lapla_problem} on the unit square (top row) and the unit cube (bottom row). }\label{fig:23d_error_curve}
%\end{figure}

\begin{figure}[!htbp]
\centering
{\includegraphics[width=5.5cm]{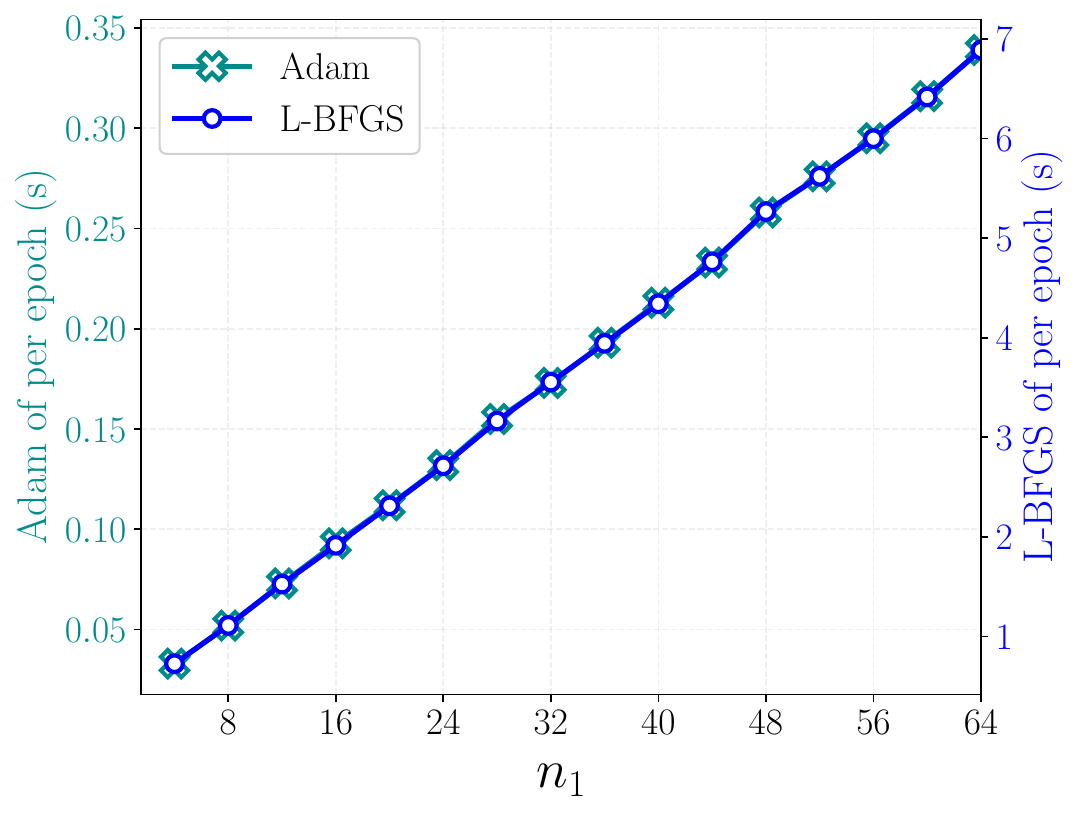}}\hspace{1cm}
{\includegraphics[width=5.5cm]{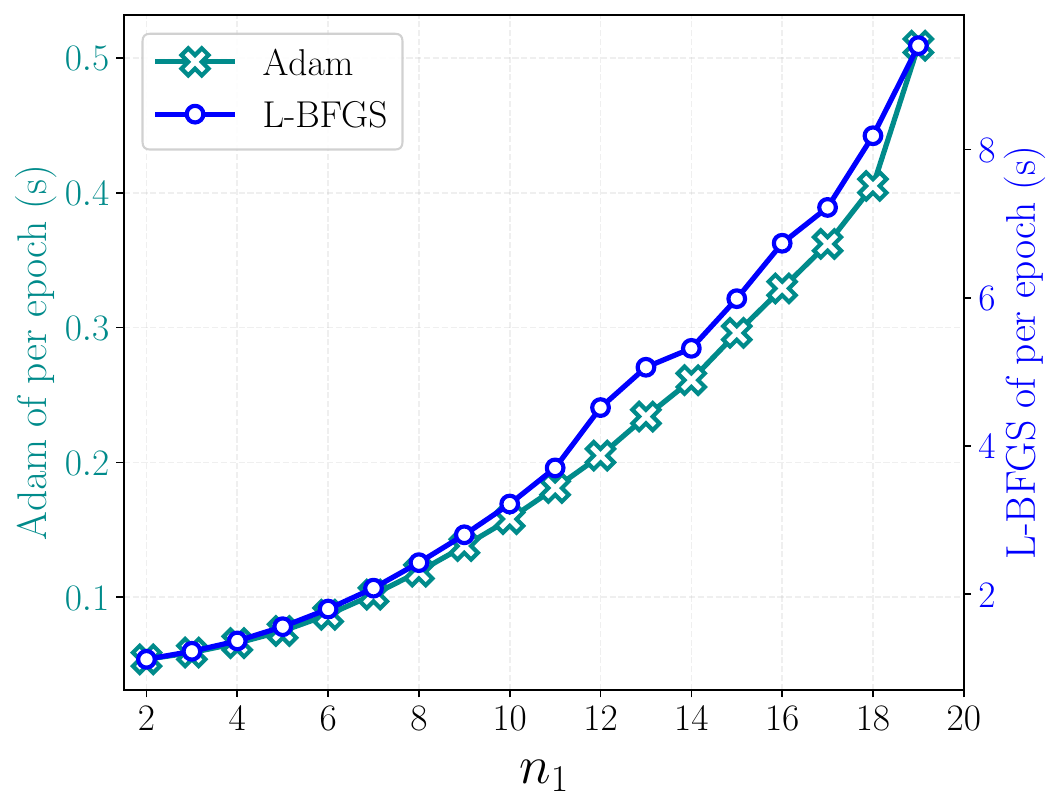}}
\caption{The average computational time per epoch of fTNN for solving \cref{eq:fra_lapla_problem} with direction resolution $n_{1}$ on the unit square (left) and the unit cube (right). }
\label{fig:23d_dif_dir_error}
\end{figure}

We also examine how the accuracy-cost trade-off depends on the directional resolution \(n_{1}\). The average computational time per epoch of fTNN is drawn against $n_{1}$ in \cref{fig:23d_dif_dir_error}, which shows a deterministic trend. Relative $L^{2}$ errors in both two and three dimensional cases are shown in \cref{tab:2d_fpe_n1,tab:3d_fpe_n1}. The observed error reduction is consistent with the \(O(n_1^{-2})\) decay predicted by the 
spherical integration theory in \cref{sec:FL_Analysis}, confirming that the angular resolution dominates accuracy once the radial singularities are properly resolved. 
This trend is precisely the one targeted by the present paper: after the geometry-adaptive 
decomposition has removed the difficulty caused by radial singularity, the further improvement comes from refining a deterministic angular discretization rather than from increasing the number of Monte Carlo samples as used in QE-MC-fPINN.

\begin{table}[!htbp]\small
\centering
\caption{Relative $L^{2}$ errors and time of fTNN solving the 2D fPE with different $n_{1}$.}\label{tab:2d_fpe_n1}
\begin{tabular}{ccccccc}
\toprule
&$(\alpha, \beta_{1}, \beta_{2})$ & $n_{1}=4$ & $n_{1}=8$ & $n_{1}=16$ & $n_{1}=32$ & $n_{1}=64$ \\
\midrule
\multirow{3}{*}{$e_{L^{2}}$}
& $(1.80,1.90,1.95)$ & 5.130e-4 & 6.639e-5 & 2.908e-5 & 2.446e-5 & 1.257e-5 \\
& $(0.80,0.90,1.20)$ & 3.545e-3 & 3.696e-4 & 1.273e-4 & 7.308e-5 & 2.297e-5 \\
& $(1.60,1.50,1.79)$ & 3.545e-3 & 3.696e-4 & 1.273e-4 & 7.308e-5 & 2.297e-5 \\
\midrule
\multirow{2}{*}{Time (s)}
& $t_1$ & 0.033 & 0.052 & 0.093 & 0.173 & 0.339 \\
& $t_2$ & 1.111 & 1.111 & 1.915 & 3.554 & 6.888 \\
%& $T$   & 255.2 & 274.2 & 476.0 & 883.8 & 1716.6 \\
\bottomrule
\multicolumn{7}{l}{\footnotesize $t_1$/$t_2$ is the average time per epoch of Adam/L-BFGS.}%; $T$ is the total training time
\end{tabular}
\end{table}

\begin{table}[!htbp] \small
\centering
\caption{Relative $L^{2}$ errors and time of fTNN solving the 3D fPE with different $n_{1}$.}\label{tab:3d_fpe_n1}
\begin{tabular}{cccccc}
\toprule
& $(\alpha, \beta_{1}, \beta_{2})$& $n_{1}=2$ & $n_{1}=4$ & $n_{1}=8$ & $n_{1}=16$ \\
\midrule
\multirow{3}{*}{$e_{L^{2}}$}
& $(1.60,1.20,1.75)$ & 7.970e-2 & 9.383e-3 & 4.268e-3 & 3.369e-3 \\
& $(1.96,1.65,2.45)$ & 1.523e-2 & 1.317e-3 & 3.590e-4 & 2.333e-4 \\
& $(1.70,1.86,2.88)$ & 4.431e-2 & 7.411e-3 & 1.850e-3 & 3.635e-4 \\
\midrule
\multirow{2}{*}{Time (s)}
& $t_1$ & 0.054 & 0.066 & 0.119 & 0.329 \\
& $t_2$ & 1.122 & 1.370 & 2.426 & 6.735 \\
%& $T$   & 278.4 & 340.0 & 604.2 & 1676.0 \\
\bottomrule
\end{tabular}
\end{table}

\begin{figure}[!tbhp]\small
\centering
\subfloat[$\alpha = 0.40$]{\includegraphics[width=3.8cm]{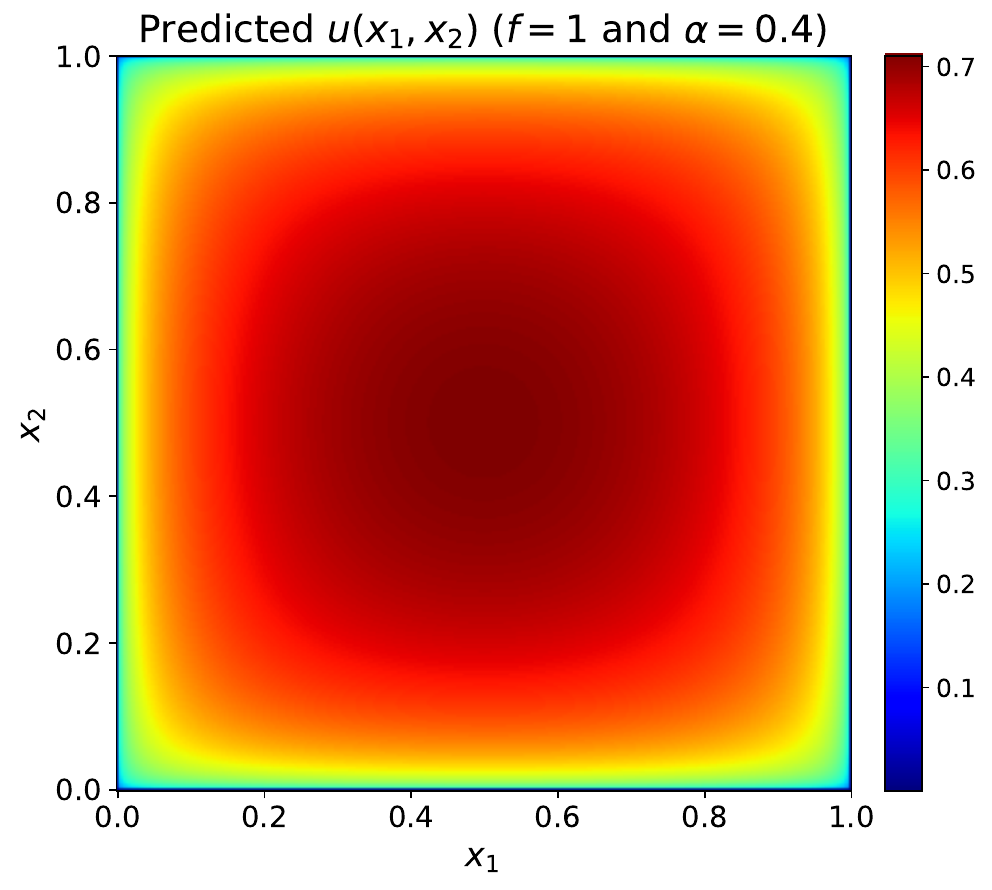}}
\subfloat[$\alpha = 0.60$]{\includegraphics[width=3.8cm]{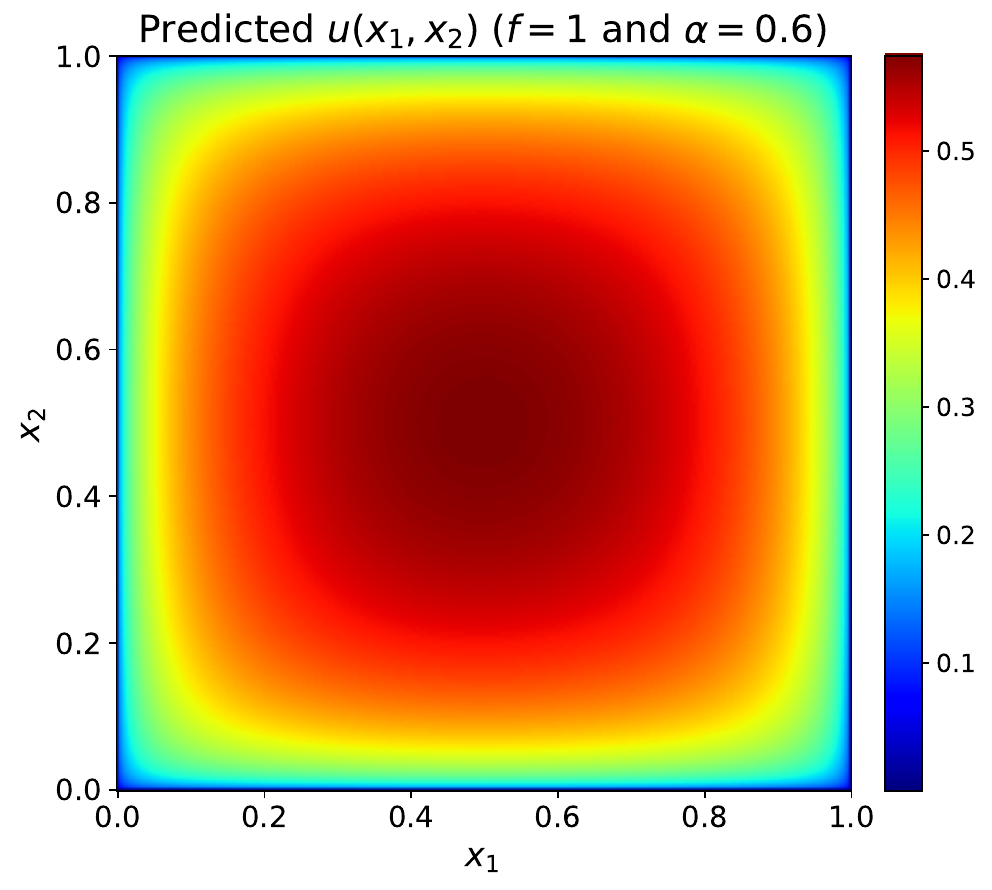}}
\subfloat[$\alpha = 1.20$]{\includegraphics[width=3.8cm]{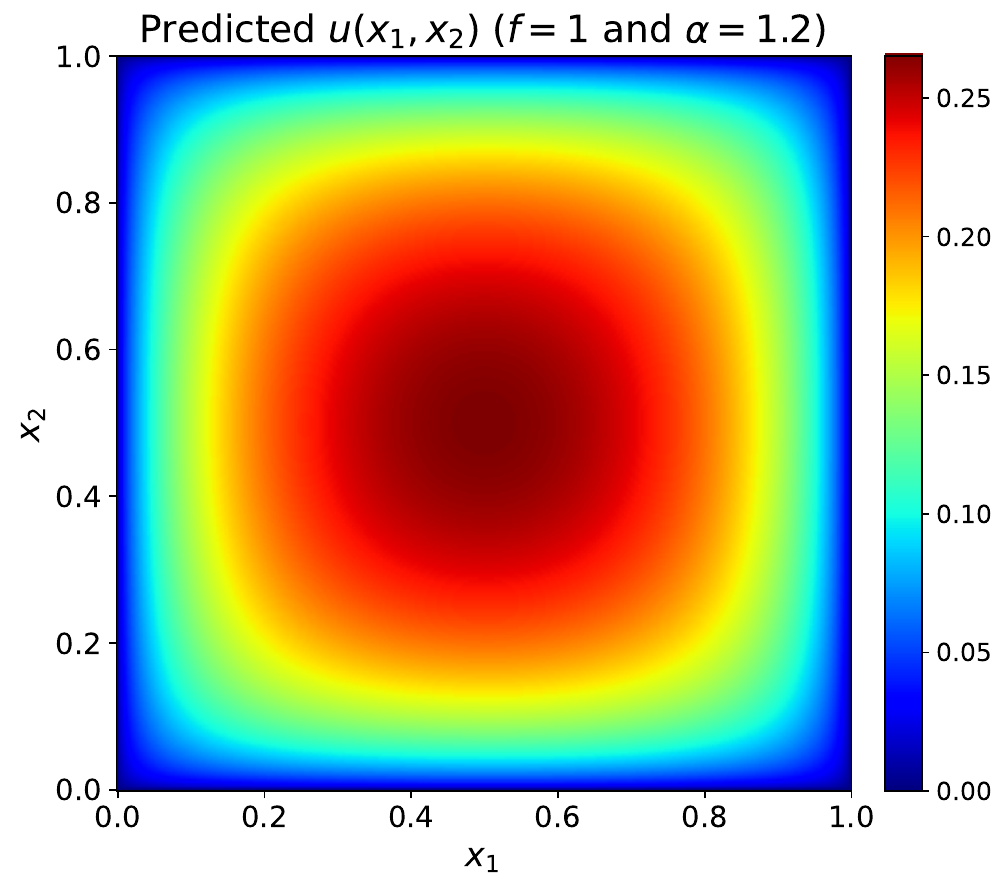}}
\subfloat[$\alpha = 1.80$]{\includegraphics[width=3.8cm]{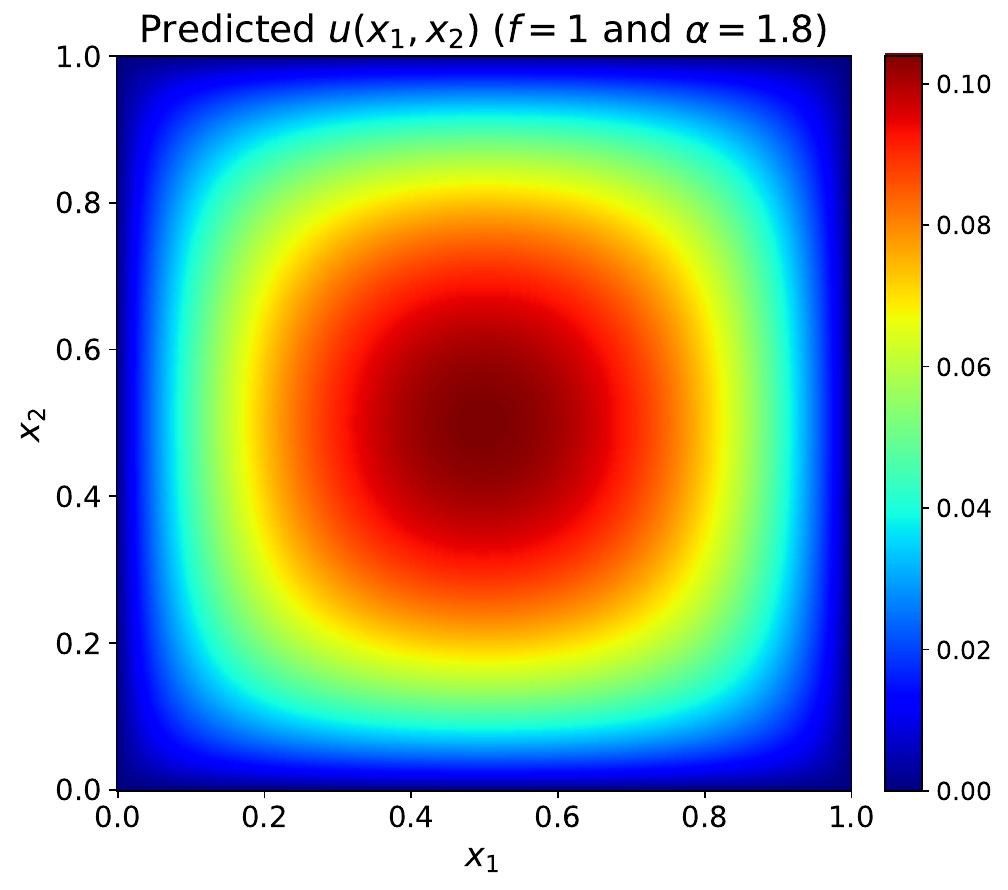}}
\caption{Solutions of fTNN to solve the 2D fPE with RHS function $f=1$.}\label{fig:2d_nosol}
\end{figure}

\begin{figure}[!htbp]
\centering
\includegraphics[width=12cm,height=4.2cm]{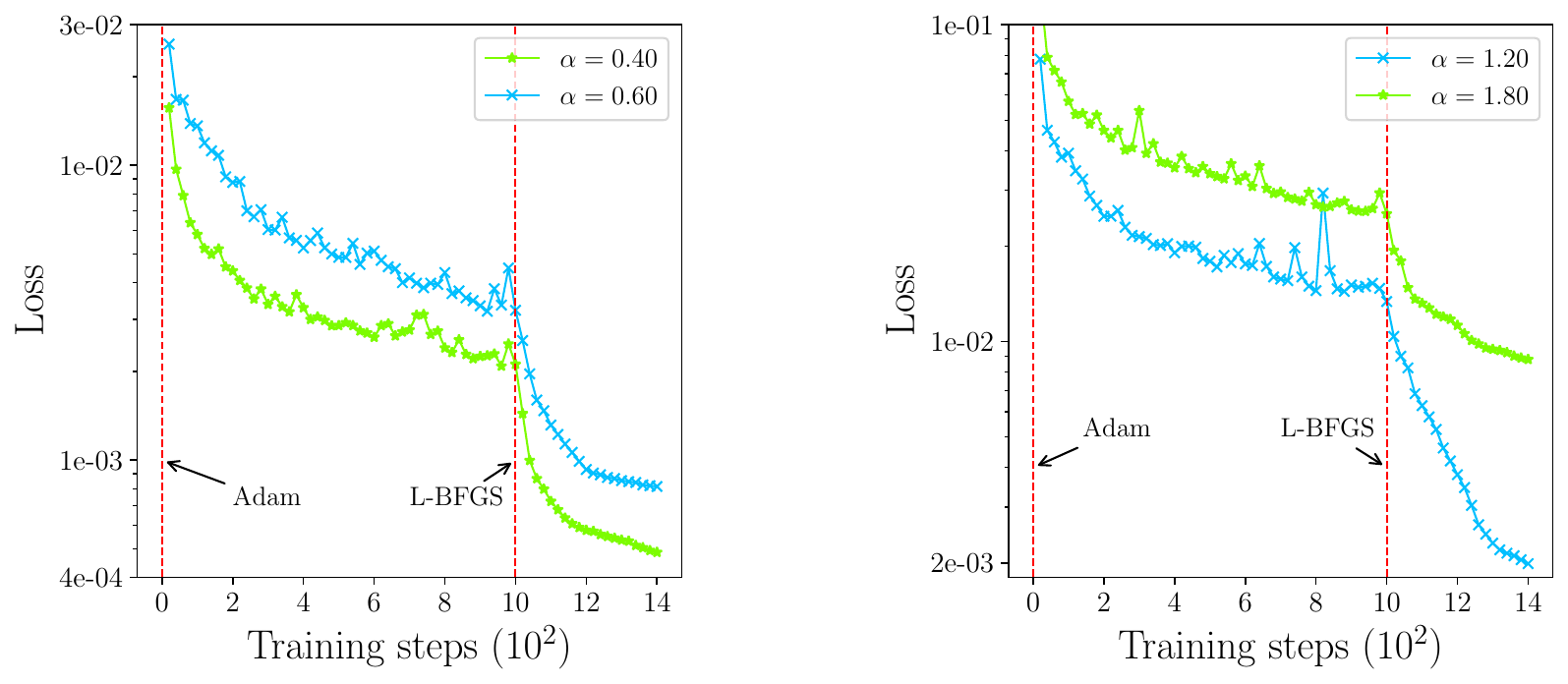}
\caption{Loss curves of fTNN to solve the 2D fPE with RHS function $f=1$.}\label{fig:2d_nonsol_loss}
\end{figure}

We then use fTNN to solve the fPE \cref{eq:fra_lapla_problem} with the RHS function $f=1$ (clearly, $s=0$) and different $\alpha$ in two dimensional case. Since the exact solutions are not known, we show the numerical solutions in \cref{fig:2d_nosol} and corresponding loss curves in \cref{fig:2d_nonsol_loss}.

\subsubsection{FPEs on the unit ball}\label{sec:unit_fPE_num}
We consider fPE \cref{eq:fra_lapla_problem} on the unit ball, for which the corresponding RHS function could be analytically derived for a given solution. 
For brevity, we present the two-dimensional case, while the extension to three dimensions is analogous. In two dimensional case, it holds that
\[
(-\Delta)^{\alpha / 2} \left(1-\|\boldsymbol{x}\|_2^{2}\right)^{1+\alpha/2} = 2^{\alpha}\,\Gamma(\alpha / 2+2)\,\Gamma(\alpha / 2+1)\,\left(1-(\alpha/2+1)\|\boldsymbol{x}\|_2^{2}\right).
\]
Using polar coordinates \(\boldsymbol{x}=(\rho\cos v,\rho\sin v)\) with \(\rho\in[0,1)\) and \(v\in[0,2\pi)\), the RHS function reads
\[
f(\rho,v)=2^{\alpha}\,\Gamma(\alpha / 2+2)\,\Gamma(\alpha / 2+1)\,\left(1-(\alpha / 2+1)\rho^{2}\right).
\]
Observe that 
\begin{equation*}
\lim\limits_{\rho\to 1^{-}}f(\rho,v) = -2^{\alpha-1}\alpha\,\Gamma(\alpha / 2+2)\,\Gamma(\alpha / 2+1) \neq 0.
\end{equation*}
Consequently, the limit
\[
\lim_{\rho\to 1^{-}}\frac{f(\rho,v)}{(1-\rho^{2})^{s}}
\]
is finite and nonzero if and only if the denominator tends to a constant, i.e., \(s= 0\). Hence, we employ the BFE strategy for this problem.

The approximate solution is expanded in terms of neural network basis functions: 
\begin{equation*}
\Psi(\rho, v) = \sum\limits_{j=1}^p c_j \varphi_j(\rho, v).
\end{equation*}
Then
\begin{equation*}
(-\Delta)^{\alpha /2} \varphi_j(\rho, v) = C_{2, \alpha} \left( I_{1,j}(\rho, v) + I_{2,j}(\rho, v) \right).
\end{equation*}
The near-field integral is approximated as
\[
I_{1,j}(\rho,v) \approx  r_{0}(\rho)^{-\alpha}\sum_{l=1}^{N_0}  \sum_{k=1}^{N} w_lw_k^{(0,1-\alpha)} \frac{F_j(\rho, v, r_{0}(\rho)\tau_k^{(0,1-\alpha)}, \eta_l)}{(\tau_k^{(0,1-\alpha)})^{2}},
\]
where $r_{0}(\rho)=1-\rho$, $F_j(\rho, v, r, \eta) 
= 2\varphi_j(\rho, v) - \varphi_j(\rho', v') - \varphi_j(\rho'', v'')$, with
\begin{equation*}
\rho' = \sqrt{\rho^{2} + r^{2} + 2\rho r \cos(\eta - v)}, \quad \rho''  = \sqrt{\rho^{2} + r^{2} - 2\rho r \cos(\eta - v)},
\end{equation*}
and
\begin{equation*}
v'=\arctan\left( \frac{\rho \sin v + r \sin\eta}{\rho \cos v + r \cos\eta} \right), \quad v'' = \arctan\left( \frac{\rho \sin v - r \sin\eta}{\rho \cos v - r \cos\eta} \right).
\end{equation*}

The far-field integral is then approximated as
\[
I_{2,j}(\rho,v) \approx \sum_{i=1}^{2} \sum_{k=1}^{2n_{i}}\frac{\pi}{n_{i}}Q_{i,j}(\rho,v, \eta_{ik}),
\]
with $\eta_{ik}$ being discrete directional angles. $Q_{1,j}$ and $Q_{2,j}$ are defined as
\begin{equation*}
Q_{1,j}(\rho,v,\eta) \approx \left(d_{\boldsymbol{x}}(\eta)-r_{0}(\rho,v)\right) \sum_{m=1}^{N_0} w_m\frac{\varphi_j(\rho,v) - \varphi_j(\rho',v')}{\left(r(\eta,t_m)\right)^{1+\alpha}},\quad
Q_{2,j}(\rho,v,\eta) = \frac{\varphi_j(\rho,v)}{\alpha \left(d_{\boldsymbol{x}}(\eta)\right)^{\alpha}},
\end{equation*}
where \(r(\eta,t):=r_{0}(\rho,v)+\left(d_{\boldsymbol{x}}(\eta)-r_{0}(\rho,v)\right)t\) and the distance \(d_{\boldsymbol{x}}(\eta)\) from the point \((\rho,v)\) to the boundary along direction \(\eta\) is given by
\[
d_{\boldsymbol{x}}(\eta) = -\rho \cos(\eta - v) + \sqrt{1 - \rho^{2} \sin^{2}(\eta - v)}.
\]

The above formulas provide the numerical approximation of fractional Laplacian in two-dimensional polar coordinates. All computations are performed in polar coordinates, with the neural network basis functions taking polar coordinates as inputs.

The experiments on the unit ball provide a well-controlled benchmark for evaluating fTNN against several state-of-the-art methods based on neural network. The availability of analytic solutions exhibiting canonical boundary singularities, combined with the spherical geometry that admits highly accurate deterministic integration, ensures a meaningful comparison. In the following experiments, MC-fPINN, Improved MC-fPINN, and QE-MC-fPINN all employ the same PINN backbone and training protocol. The specific hyperparameter settings and implementation details are adopted from \cite{hu2024tackling}. The angular resolution $n_1$ of fTNN is set to be $32$ in 2D case and $16$ in 3D case, while fPINN is configured as described in \cite{pang2019fpinns}. 

\begin{table}[htb]\small
\centering
\caption{Relative $L^2$ test errors using different methods to solve the fPEs on the unit balls.}\label{tab:23d_ball_error} 
\begin{tabular}{clcccc}
\toprule
\multirow{2}{*}{$u_{\text{exact}}$} & \multirow{2}{*}{Method}
& \multicolumn{2}{c}{$d=2$} & \multicolumn{2}{c}{$d=3$} \\ 
\cmidrule(lr){3-4} \cmidrule(lr){5-6}
& & $\alpha=1.5$ & $\alpha=1.9$ & $\alpha=1.5$ & $\alpha=1.9$ \\
\midrule
\multirow{5}{*}{$(1-\|\boldsymbol{x}\|^2)^{1+\alpha/2}$} 
& MC-fPINN             & 1.78e-2 & 5.31e-1 & 1.37e-2 & 8.11e-1 \\
& Improved MC-fPINN    & 4.88e-3 & 4.96e-1 & 7.66e-3 & 4.35e-1 \\
& fPINN                & 3.61e-3 & 1.12e-3 & 5.02e-3 & 5.59e-3 \\
& QE-MC-fPINN          & 2.00e-3 & 1.33e-3 & 1.74e-3 & 1.18e-3  \\
& fTNN & 2.20e-6 & 1.70e-5 & 3.97e-4 & 3.56e-4 \\

\midrule
\multirow{5}{*}{$(1-\|\boldsymbol{x}\|^2)^{\alpha/2}$} 
& MC-fPINN             & 7.01e-1 & 7.20e-1 & 4.43e-1 & 8.57e-1 \\
& Improved MC-fPINN    & 4.24e-1 & 5.12e-1 & 2.65e-1 & 1.88e-1 \\
& fPINN                & 1.36e-2 & 1.34e-2 & 4.95e-1 & 1.30e-0 \\
& QE-MC-fPINN          & 3.41e-2 & 6.05e-3 & 3.07e-2 & 3.00e-2 \\
& fTNN & 2.17e-5 & 1.72e-5 & 3.74e-4 & 1.58e-4 \\
\bottomrule
\end{tabular}
\end{table}

\Cref{tab:23d_ball_error} reports the relative $L^2$ test errors $e_{\mathrm{test}}$ of four different methods for solving two- and three-dimensional fPEs on the unit balls, with exact solutions
\begin{equation}\label{uesmooth}
u_{\mathrm{exact}}=(1-\|\boldsymbol{x}\|^2)^{1+\alpha/2},
\end{equation}
and
\begin{equation}\label{uesingular}
u_{\mathrm{exact}}=(1-\|\boldsymbol{x}\|^2)^{\alpha/2},
\end{equation}
for fractional orders $\alpha=1.5$ and $\alpha=1.9$.

\begin{figure}[!tbhp]
\centering
\includegraphics[width=12cm,height=4.2cm]{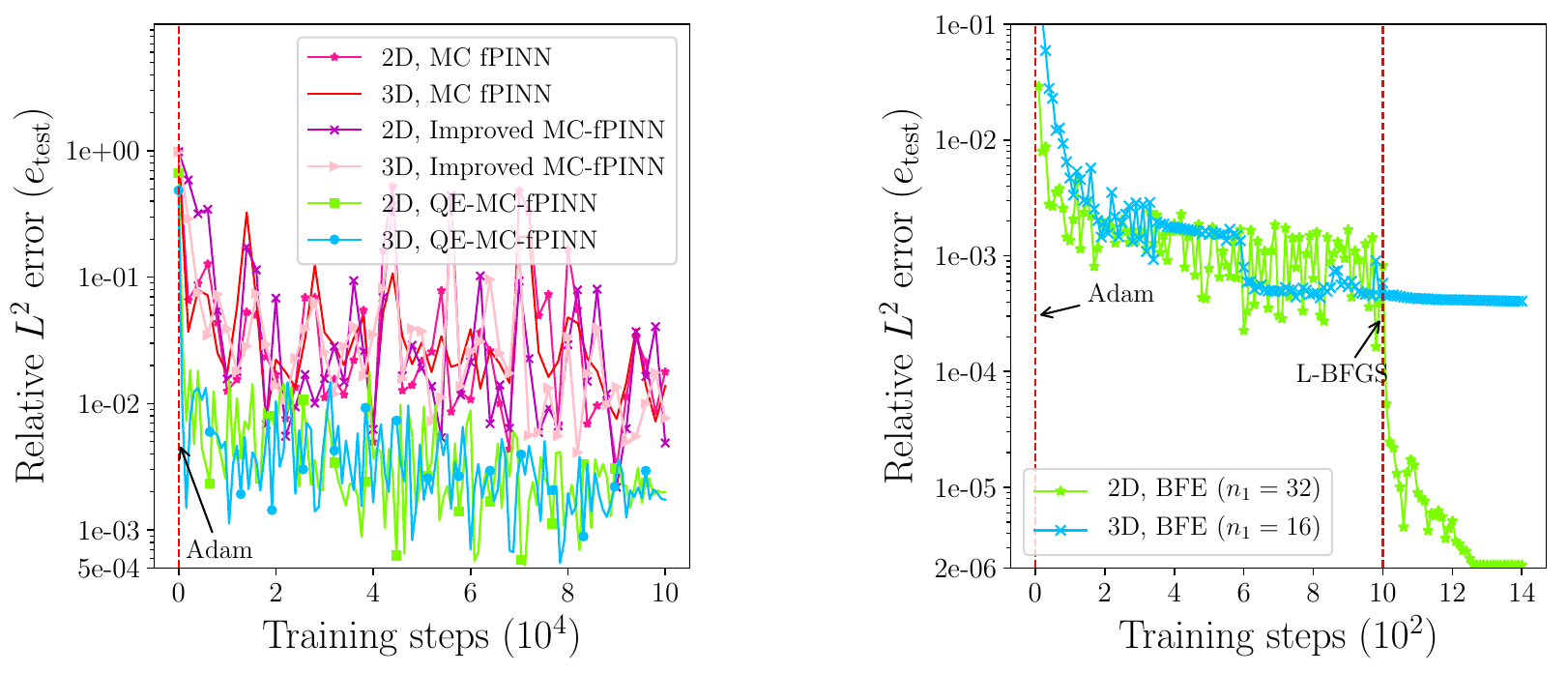}
\caption{Relative $L^{2}$ test error curves of fPINN-based methods (left column) and fTNN (right column) for solving fPE with exact solution \cref{uesmooth}, $\alpha = 1.5$.}\label{fig:c_loss_err_three_ways}
\end{figure}

For solution \cref{uesmooth} with $\alpha=1.5$, all five methods behave well, although fTNN achieves substantially lower errors than the other methods. For solution \cref{uesmooth} with $\alpha=1.9$, there are more pronounced differences among the errors of different methods. To be specific, the relative errors of MC-fPINN and Improved MC-fPINN are the greatest, fPINN and QE-MC-fPINN reduce them by about two orders of magnitude, while fTNN achieves a further reduction. For the more singular solution \cref{uesingular} in 2D case, the comparison results are similar to that for solution \cref{uesmooth} with $\alpha=1.9$. While for that in 3D case, fPINN exhibits unsatisfactory performance. Notably, fTNN achieves substantially lower errors than the other methods across all test cases. 

\begin{figure}[!htbp]
\centering
\includegraphics[width=12cm]{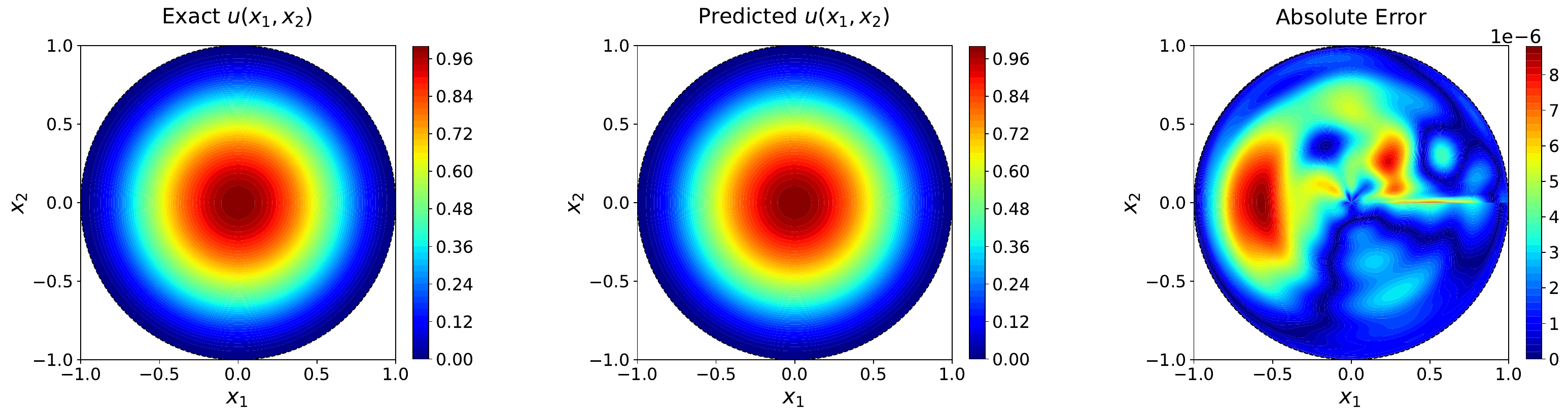}
\includegraphics[width=12cm]{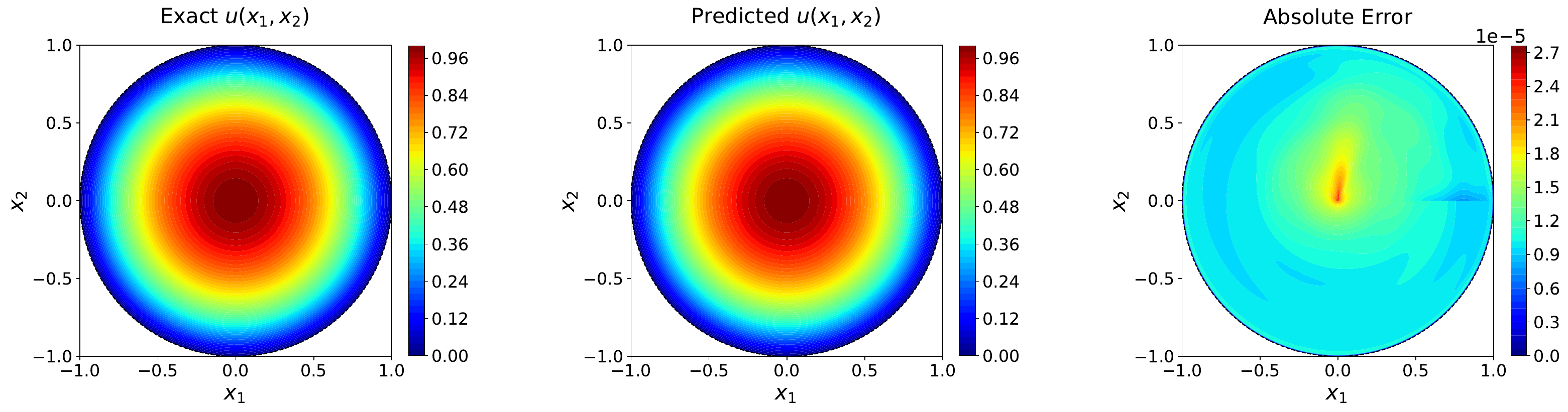}		
\caption{Plots of fTNN to solve fPE \cref{eq:fra_lapla_problem} on the unit ball. Top: exact solution \cref{uesmooth} with $\alpha=1.5$, numerical solution, and the absolute error. Bottom: exact solution
\cref{uesingular} with $\alpha = 1.9$, numerical solution, and the absolute error.} \label{fig:2d_bar_ball}
\end{figure}

The relative error curves in \cref{fig:c_loss_err_three_ways} demonstrate that fTNN converges smoothly and attains markedly lower errors compared with MC-fPINN, Improved MC-fPINN, and QE-MC-fPINN. We plot the solutions in \cref{fig:2d_bar_ball} which indicates that fTNN accurately captures the singular structure near boundary without the erratic oscillations and variance inherent in Monte Carlo-based methods. Overall, these results highlight that the combination of deterministic integration, boundary-aware trial functions, and adaptive singularity treatment confers substantial improvements in both accuracy and stability for low- and moderate-dimensional fPEs.

\subsection{Fractional advection-diffusion equation}
We next turn to time-dependent problems, for which \cref{Algorithm_fPDE} is used (This is what the fTNN refers to in this subsection). We consider the time-space fractional advection-diffusion equation (ADE) on the unit ball, with convection coefficient \(\boldsymbol{v} = (0.1,0.1)\) for $d=2$ and \(\boldsymbol{v} = (0.1,0.1,0.1)\) for $d=3$, which is a typical case of the fPDE defined in \cref{eq:fPDE}. We employ the following quadrature scheme
\begin{align*}
\Gamma(1-\gamma)\cdot{}_0^C D_t^{\gamma}\left( t^{\gamma}\phi_{t,j}(t) \right) 
= \int_0^1 (1-\tau)^{-\gamma} \tau^{\gamma-1} S_{1,j}(t,\tau) \mathrm{d}\tau 
\approx \sum_{k=1}^{N} w_k^{(-\gamma, \gamma-1)} S_{1,j}\left(t,\tau_k^{(-\gamma, \gamma-1)}\right),
\end{align*}
for the time-fractional Caputo derivative, where 
\begin{equation*}
S_{1,j}(t,\tau) = \gamma\phi_{t,j}(t\tau) + t\tau \phi_{t,j}'(t\tau),
\end{equation*}
see \cite{lin2025TFPIDE} for further details. 

For the spatial operator, we use the same deterministic discretization as in \cref{sec:unit_fPE_num}; the new feature in this subsection is that this spatial solver is coupled with the spatiotemporally separable subspace representation and alternating optimization. In the two-dimensional case, we define the trial function \(\Psi\) in \cref{eq:trai_fun} as
\begin{equation}\label{ADE_trial}
\Psi(\rho,\eta,t; c, \theta) = \widehat{\Psi}(\rho,\eta,t;c,\theta)+1-\rho^{2},
\end{equation}
with
\begin{equation*}
\widehat{\Psi}(\rho,\eta,t;c,\theta) = \sum\limits_{j=1}^{p}c_j t^{\gamma}\phi_{t,j}(t) (1-\rho^{2})^{\mu_{j}}\phi_{1,j}(\rho)\phi_{2,j}(\eta).
\end{equation*}

%\begin{figure}[!tbhp]
%\centering
%\includegraphics[width=14cm]{fig/2D_STfADE_car_s_0.75_T_1_t0_0.51.pdf}	
%\includegraphics[width=14cm]{fig/3D_STfADE_s_0.75_T_1_t0_0.52.pdf}
%\caption{Numerical results of fTNN for solving the time-space fractional ADE ($\gamma=0.5$, $\alpha=1.50$). Top row: 2D, $n_{1}=32$; bottom row: 3D, $n_{1}=16$: the exact solutions (left column), the solutions of fTNN (middle column), and the absolute errors between them (right column).} \label{fig:STfPDE}
%\end{figure} 

We first examine the performance of the new method in short-time simulations. 
The time interval $[0,1]$ is divided into $4$ subintervals, with $10$ Gauss quadrature points selected within each subinterval to ensure sufficient accuracy for both two- and three-dimensional time-space fractional ADEs. 
The exact solution is taken as 
\begin{equation*}
u_{\text{exact}}(\boldsymbol{x},t) = e^{-t}\left(1-\|\boldsymbol{x}\|_2^{2}\right)^{1+\alpha /2}.
\end{equation*}
The corresponding source term \(f(\boldsymbol{x},t)\) for both two and three dimensions is provided in \cite{pang2019fpinns}.

%\begin{table}[!tbhp]\small
%\centering
%\caption{Relative $L^{2}$ errors $e_{\mathrm{test}}$ of the time-space fractional ADEs using fPINN and fTNN.}\label{tab:fPDE_fPINN_BFE_compact}
%\begin{tabular}{ccc}
%\toprule
%Problem & fPINN & fTNN  ($n_{1}=8$)\\
%\midrule
%2D, $\gamma = 1.0$ & 1.066e-3 & 6.233e-6  \\
%2D, $\gamma = 0.5$ & 1.241e-3 & 6.130e-5  \\
%3D, $\gamma = 1.0$ & 2.359e-3 & 4.110e-4  \\
%3D, $\gamma = 0.5$ & 2.758e-3 & 7.942e-4  \\
%\bottomrule
%\end{tabular}
%\end{table}

\begin{table}[!tbhp]\small
\centering
\caption{Relative $L^{2}$ test errors of the time-space fractional ADEs using fPINN and fTNN.}\label{tab:fPDE_fPINN_BFE_compact}
\begin{tabular}{cccccc}
\toprule
\multirow{2}{*}{Problem} & \multirow{2}{*}{fPINN} & \multicolumn{4}{c}{fTNN}\\
\cmidrule{3-6}
& & $n_{1}=2$ & $n_{1}=4$ & $n_{1}=8$& $n_{1}=16$ \\
\midrule
2D, $\gamma = 1.0$ & 1.066e-3 & 2.051e-4 & 1.803e-5 & 6.233e-6 & 7.217e-6  \\
2D, $\gamma = 0.5$ & 1.241e-3 & 1.041e-3 & 6.891e-5 & 6.130e-5 & 7.408e-5  \\
3D, $\gamma = 1.0$ & 2.359e-3 & 2.294e-2 & 3.402e-3 & 4.110e-4 & 3.158e-4   \\
3D, $\gamma = 0.5$ & 2.758e-3 & 2.302e-2 & 3.444e-3 & 7.942e-4 & 2.102e-4   \\
\bottomrule
\end{tabular}
\end{table}

\begin{figure}[!tbhp]
\centering
\includegraphics[width=12cm,height=4.2cm]{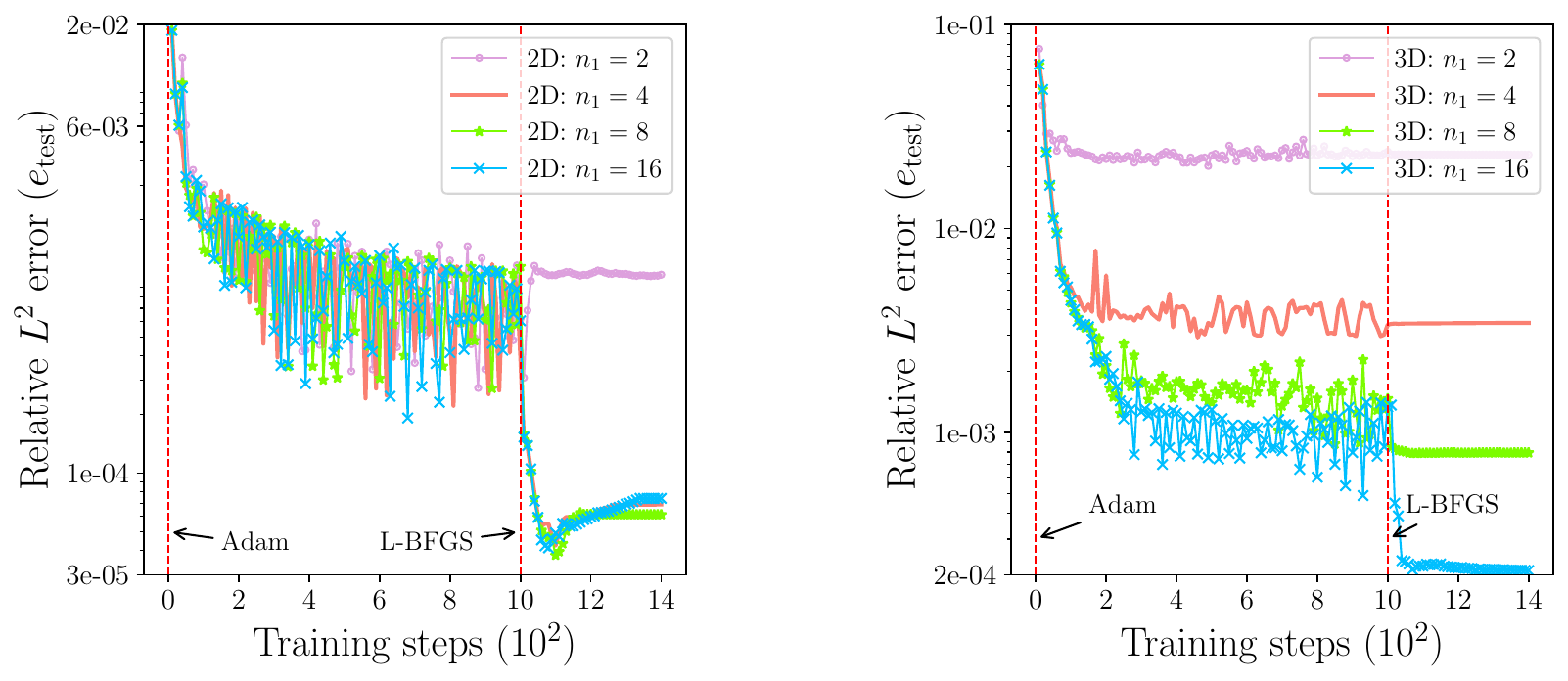}
\caption{Relative $L^{2}$ test errors for solving time-space fractional ADE ($\gamma=0.5,  \alpha=1.50$). }
\label{fig:23d_STfPDE_error}
\end{figure}

Substituting the decomposition \cref{ADE_trial} into \cref{eq:fPDE} transforms the original problem into a homogeneous initial-boundary value problem for \(\widehat{\Psi}\), with the RHS function given by
\begin{align*}
\widehat{f} = &2^{\alpha}\Gamma(\alpha/2+2)\Gamma(\alpha /2+1)(e^{-t}-1)\left(1-(\alpha /2+1)\rho^{2}\right)
-(1-\rho^{2})^{1+\alpha /2}t^{1-\gamma}E_{1,2-\gamma}(-t) \\
&- 0.2\rho(\alpha /2+1)(e^{-t}-1)(1-\rho^{2})^{\alpha /2}(\cos\eta+\sin\eta),
\end{align*}
where \(E_{\cdot, \cdot}(z)\) is the Mittag-Leffler function.

The relative $L^{2}$ test errors of different methods are shown in \cref{tab:fPDE_fPINN_BFE_compact}. As seen in \cref{tab:23d_ball_error}, fPINN achieves lower error than MC-fPINN and Improved MC-fPINN methods, here we list just fPINN among those fPINN-based methods to prevent redundancy. Relative to fPINN, fTNN reduces the errors to the \(10^{-5}\) level in 2D and to the \(10^{-4}\) level in 3D. The relative error curves in \cref{fig:23d_STfPDE_error} reveal that fTNN exhibits smooth convergence histories. These results highlight the benefit of coupling deterministic spatial quadrature with the STSNN subspace formulation. 
\begin{table}[!htbp]\small
\centering
\caption{Relative $L^{2}$ test errors of fTNN for solving time-space fractional ADEs and average per-epoch times (seconds) of Adam and L-BFGS for long-time simulations.}
\label{tab:fPDE_long_time}
\begin{tabular}{ccccccc}
\toprule
Case&$(\alpha , \gamma, \gamma_{1}, \gamma_{2})$&$T=50$&$T=100$&$T=150$ & Adam & L-BFGS\\
\midrule
2D, case A1&$(1.90,0.85,0.85,1.70)$ & 3.363e-5  & 9.074e-5 &3.405e-5 & 0.119s & 0.744s\\
2D, case A2&$(1.30,0.45,0.75,0.80)$ & 2.558e-4  & 1.889e-4 &1.353e-4 & 0.121s & 0.758s \\
3D, case B1&$(1.90,0.85,0.85,1.00)$ & 9.140e-5  & 9.486e-5 &8.933e-5 & 0.413s & 8.445s\\
3D, case B2&$(1.30,0.45,0.75,0.80)$ & 4.258e-4  & 4.376e-4 &4.411e-4 & 0.419s & 8.544s\\
\bottomrule
\end{tabular}
\end{table}

We then proceed to discuss the case of long-time simulations. Since long-time simulation of fPDEs faces far more severe theoretical and computational obstacles than short-time counterpart, the performance of fTNN deserves focused and in-depth discussion.
To this end, the exact solution is set as 
\begin{equation}\label{uexact_long}
u_{\text{exact}}(\boldsymbol{x},t)=(t^{\gamma_{2}}+t^{\gamma_{1}})(1-\|\boldsymbol{x}\|_2^{2})^{1+\alpha /2}.
\end{equation}
During the computation of the loss function, the time interval \([0,T]\) is divided into 25 subintervals, with $16$ Gauss quadrature points selected within each subinterval. As shown in \cref{tab:fPDE_long_time}, the relative \(L^{2}\) test errors at final times \(T=50,100,150\) remain in the \(10^{-4}\)-\(10^{-5}\) range, indicating that the chosen temporal quadrature is sufficiently dense to resolve the memory term even over long horizons. 

\begin{table}[!htbp]\small
\centering
\caption{Relative $L^{2}$ test errors using different methods (long-time simulation, $T=100$).}
\label{tab:comparison_T100}
\begin{tabular}{cccccc}
\toprule
Case  &  MC-fPINN  & Improved MC-fPINN &fPINN     & QE-MC-fPINN & fTNN \\                                                       
\midrule                                                 
2D, case A2 &  1.001e-01 & 2.321e-01  & 4.066e-02 & 6.208e-03 & 1.889e-4 \\
3D, case B1 &  9.851e-01 & 3.717e-01  & 3.371e-03 & 1.222e-03 & 9.486e-5 \\
\bottomrule
\end{tabular}
\end{table}
We also compare the relative $L^{2}$ test errors at $T=100$ (for case A2 and case B1 in \cref{tab:fPDE_long_time}) of different methods in \cref{tab:comparison_T100}. 
The fPINN results are obtained using the DeepXDE implementation \cite{lu2021deepxde} with the $L1$ scheme and Gr\"unwald--Letnikov discretization configured as in \cite{pang2019fpinns}. The MC-fPINN, Improved MC-fPINN, and QE-MC-fPINN use the same hyperparameters as reported in \cite{ma2026quadrature}. All other solver configurations are kept the same as in their original references. The comparison result is similar to that in \cref{sec:unit_fPE_num}: the relative errors of MC-fPINN and Improved MC-fPINN are the greatest, fPINN and QE-MC-fPINN reduce them by one or two orders of magnitude, while fTNN achieves a further reduction in error. These results further verify the advantages of STSNN subspace formulation for temporally nonlocal dynamics, which is the third contribution of this paper. 

%These results demonstrate that the three key ingredients of the proposed framework-fully deterministic angular quadrature, boundary-singularity-aware trial spaces, and STSNN subspace optimization-are effective in the long-time regime, and that the accuracy gain over existing methods is pronounced at large terminal times.

%\begin{table}[!htbp]\small
%\centering
%\caption{$L^{2}$ relative errors $e_{\mathrm{test}}$ of fTNN with $n_1=16$ solving long-time time-space fractional ADE and average per-epoch times (seconds) for Adam and L-BFGS.}\label{tab:fPDE_long_time}
%\begin{tabular}{ccccccc}
%\toprule
%Case&$(\alpha, \gamma, \gamma_{1}, \gamma_{2})$&$T=50$&$T=100$&$T=150$ & Adam & L-BFGS\\
%\midrule
%2D, case A1 & $(1.90,0.85,0.85,1.70)$ & 6.413e-5  & 6.413e-5 &8.824e-5 & 0.020 & 0.520\\
%2D, case A2 & $(1.30,0.45,0.75,0.80)$ & 2.772e-4  & 2.238e-4 &1.852e-4 & 0.019 & 0.508\\
%3D, case B1 & $(1.90,0.85,0.85,1.00)$ & 9.140e-5  & 9.486e-5 &8.933e-5 & 0.413 & 8.445\\
%3D, case B2 & $(1.30,0.45,0.75,0.80)$ & 4.258e-4  & 4.376e-4 &4.411e-4 & 0.419 & 8.544\\
%\bottomrule
%\end{tabular}
%\end{table}

We may further expound upon the function that STSNN serves in fTNN for solving long-time fractional PDEs. The STSNN decouples the spatial and temporal dimensions, reducing the evaluation of high-dimensional time-space fractional integrals to lower-dimensional temporal and spatial fractional integrals. For long-time simulations, this allows the use of a dense set of Gauss quadrature points in the time direction without causing prohibitive memory growth. This is precisely the aspect that distinguishes the present work most clearly from QE-MC-fPINN: beyond improving the spatial operator evaluation, it furnishes a practical subspace mechanism for accurate long-time simulation of time-space fractional ADEs on bounded domains. The per-epoch costs reported in \cref{tab:fPDE_long_time} remain moderate in 2D and still manageable in 3D, which shows that the deterministic/subspace reformulation improves accuracy without destroying computational feasibility on structured domains.

\begin{figure}[!tbhp]
\centering
\includegraphics[width=12cm,height=4.2cm]{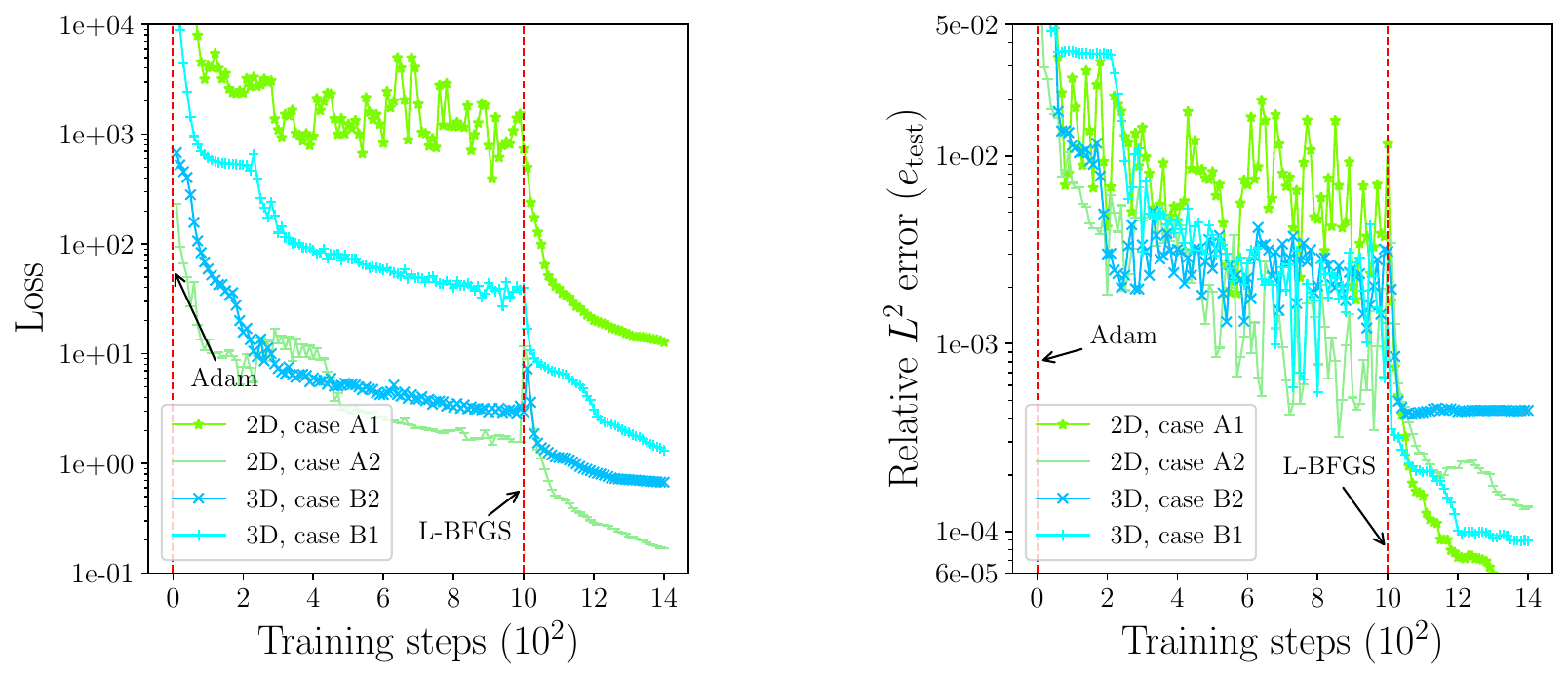}
\caption{The loss curve and the relative $L^{2}$ test error of fTNN for solving time-space fractional ADE (long-time simulation, $T=150$).}
\label{fig:LSTfPDE_loss_error}
\end{figure}

The loss curves in \cref{fig:LSTfPDE_loss_error} further indicate that the alternating STSNN subspace optimization is stable in the long-time regime: the relative \(L^{2}\) test errors reach the \(10^{-4}\) level within the first \(100\) epochs and then settle after roughly \(200\) epochs. At the final horizon \(T=150\), this corresponds to about \(430\,\mathrm{s}\) total runtime in 2D and about \(2100\,\mathrm{s}\) in 3D, which is acceptable given the difficulty of the nonlocal time-space operator.

\section{Conclusion}
We have developed a fully deterministic, subspace-based neural framework for fractional PDEs on  bounded domains with structured geometries up to three dimensions. It rests on three mutually reinforcing components: (i) geometry-adapted quadrature that resolves singular radial integrals with Gauss-Jacobi rules and replaces Monte Carlo sampling by deterministic angular quadrature; (ii) boundary-singularity-aware trial spaces whose leading exponents are adaptively selected (BFE/BRFE) to match the asymptotic structure induced by the operator and the source term; and (iii) a spatiotemporally separable neural network with alternating subspace optimization that factorizes the time-space residual into lower-dimensional integrals.

Numerical experiments from one to three dimensions show that the method yields accurate and stable approximations, with the largest gains in regimes of strong boundary singularities and long-time simulations-precisely where existing neural network solvers struggle most. A systematic study of the angular resolution confirms that, after the radial singularity is removed, the error decays consistently with the \(O(n_1^{-2})\) rate predicted by spherical integration theory, making deterministic angular quadrature the primary accuracy lever without Monte Carlo noise.

More broadly, this framework demonstrates that on structured domains, a careful synthesis of problem-adapted singular quadrature, explicit boundary enrichment, and separable tensor-network architecture can simultaneously eliminate stochastic noise and mitigate ill-conditioning-two longstanding obstacles in neural network-based fractional PDE solvers.

Future work will extend the approach to more general geometries, nonlinear fractional models, and hybrid deterministic-stochastic strategies that preserve the present accuracy while improving scalability to higher dimensions.

%\section*{Data availability}
%No external experimental data were used in this study. The numerical data, source code, and scripts that support the findings of this work are available from the corresponding author upon reasonable request.

%========================================================

\bibliographystyle{plainurl}
\bibliography{ref}

\end{document}